\documentclass{article} 
\usepackage{times}
\newif\ifarxiv
\newif\ifcameraready

\camerareadyfalse
\arxivtrue

\ifcameraready
    \usepackage{iclr2027_conference}
    \iclrfinalcopy
\else
    \ifarxiv
        \usepackage{iclr2027_arxiv}
        \iclrfinalcopy
    \else
        \usepackage{iclr2027_conference}
    \fi
\fi

\usepackage{amsmath,amsfonts,bm}

\def\eqref#1{equation~\ref{#1}}

\def\1{\bm{1}}

\DeclareMathAlphabet{\mathsfit}{\encodingdefault}{\sfdefault}{m}{sl}
\SetMathAlphabet{\mathsfit}{bold}{\encodingdefault}{\sfdefault}{bx}{n}

\usepackage{hyperref}
\usepackage{url}

\usepackage{nicefrac}       
\usepackage[table]{xcolor}         
\usepackage{makecell}
\usepackage{graphicx}
\usepackage{subcaption}
\usepackage{booktabs}
\usepackage{hyperref}
\usepackage{amsfonts}       
\usepackage{amsmath}
\usepackage{amssymb}
\usepackage{amsthm}
\usepackage{mathtools}
\usepackage{multirow}
\usepackage{placeins}
\usepackage{float}
\usepackage{thm-restate}
\usepackage{tabularx}
\usepackage{array}
\usepackage[capitalize,noabbrev]{cleveref}

\usepackage{algorithm}
\usepackage{algpseudocode}
\newcommand{\Rcomment}[1]{\Comment{\textcolor{gray!60}{#1}}}
\newcommand{\Lcomment}[1]{\Statex \textcolor{gray!60}{\(\triangleright\) #1}}

\newcommand{\FA}{\mathcal{F}_G}
\newcommand{\GLang}{\mathcal{A}}
\newcommand{\Lex}{\Lambda}

\newcolumntype{Y}{>{\raggedright\arraybackslash}X}
\newcommand{\CCG}{\cellcolor{gray!15}}
\def\ours{EPIC}
\def\mask{\texttt{[MASK]}}
\def\Valid{\textsc{Valid}}

\theoremstyle{plain}

\theoremstyle{definition}

\theoremstyle{remark}

\usepackage{tikz}
\DeclareRobustCommand{\circled}[1]{%
  \tikz[baseline=(char.base)]{%
    \node[shape=circle,draw,inner sep=0.3pt] (char) {#1};%
  }%
}

\title{Efficient and Parallel Inference under CFG Constraints for Diffusion Language Models}

\author{Hyundong Jin, Yo-Sub Han\\
Department of Computer Science\\
Yonsei University, Seoul, Republic of Korea \\
\texttt{\{tuzi04,emmous\}@yonsei.ac.kr}
}

\begin{document}

\maketitle

\begin{abstract}
Controlling language model outputs is essential 
for ensuring structural validity, reliability, and downstream usability, 
and diffusion language models are no exception.
Recent advances in diffusion language model decoding have extended output control 
beyond regular constraints to context-free grammar~(CFG) constraints. 
However, the prior CFG-constrained decoder can be 
up to four times slower than unconstrained decoding,
forcing a trade-off between the correctness benefits of CFG constraints 
and the decoding efficiency of diffusion models.
A key source of this overhead is sequential validity checking, 
which limits parallel token commitment and adds repeated validation costs.
We propose \ours{}, a CFG-constrained decoding framework designed to
improve this efficiency-correctness trade-off. 
In order to reduce decoding overhead without sacrificing
syntactic correctness, \ours{} combines lexing memoization, 
relaxed compatible subset selection for parallel commit, 
and validation using Earley-style parsing instead of deterministic automata.
This design enables multiple compatible tokens to be committed together 
while avoiding repeated lexing and expensive validation.
Experiments on three benchmarks using four models show that 
\ours{} improves the efficiency-correctness trade-off, 
bringing runtime close to unconstrained levels
while maintaining comparable syntactic and functional correctness.
Relative to the prior CFG-constrained decoder, 
\ours{} achieves a best-case inference-time reduction of 67.2\%.
Our implementation is
available at \url{https://github.com/hyundong98/EPIC-Decoding}.
\end{abstract}

\section{Introduction}
\label{EPIC:main:sec:introduction}

Driven by the success of diffusion models in vision,
there has been growing interest in applying diffusion-based methods
to language modeling~\citep{HoJA2020,SahooASGMCRK2024,NieZYZOHZLWL2025}.
Diffusion language models~(DLMs) generate text through iterative denoising
rather than autoregressive~(AR) next-token prediction, 
and masked diffusion language models~(MDLMs)
instantiate this paradigm by repeatedly refining partially masked sequences.
Recent progress has shown that MDLMs can scale to large language models~(LLMs), 
making DLMs a promising alternative to 
AR generation~\citep{SahooASGMCRK2024, NieZYZOHZLWL2025}.
Unlike AR models, which generate tokens strictly from left to right,
DLMs can update multiple positions in parallel, offering a distinctive
latency advantage~\citep{KimJJN2026}.

\begin{figure}
    \centering
    \includegraphics[width=0.85\linewidth]{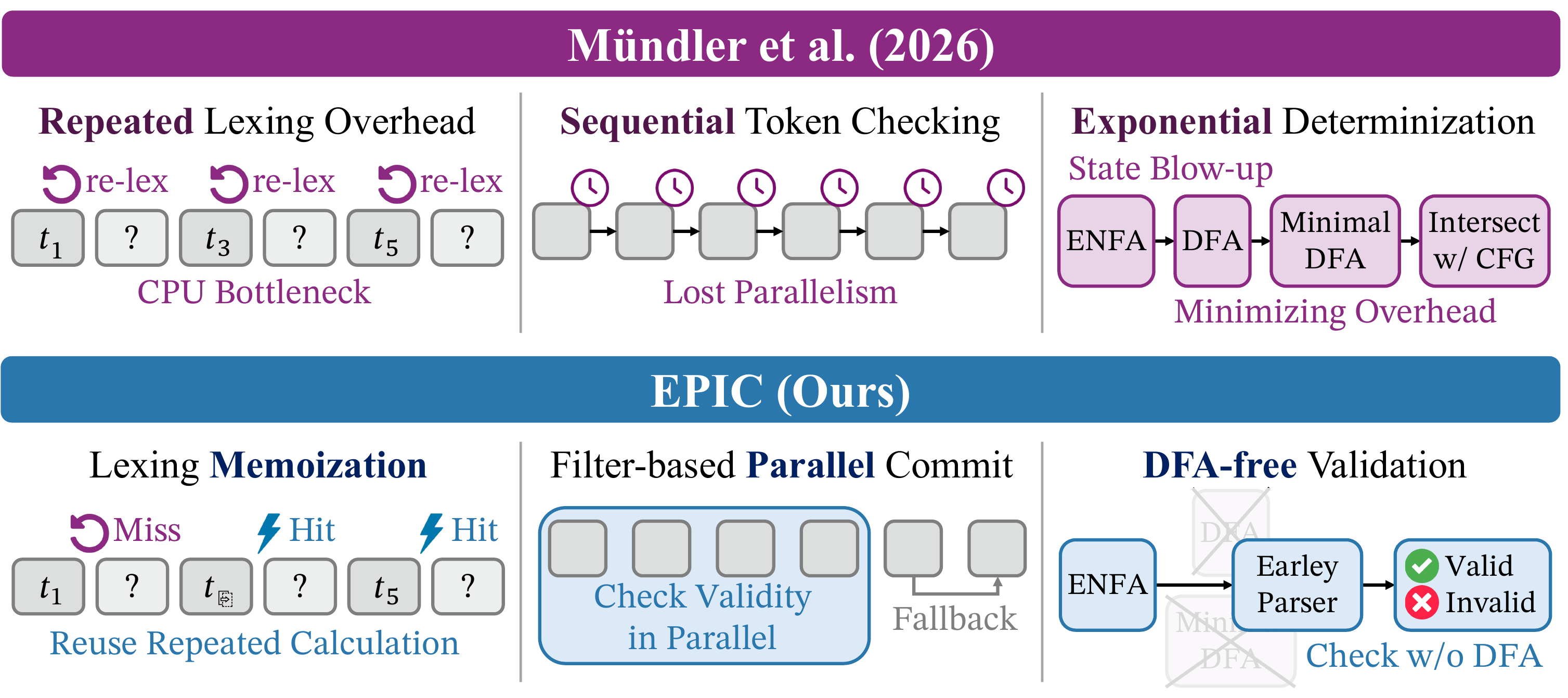}
    \caption{
    Overview of the bottlenecks in rejection-sampling-based CFG-constrained diffusion decoding and the corresponding components of \ours{}.
    }
    \label{EPIC:main:fig:moti}
\end{figure}

Despite this advantage, unconstrained DLM decoding remains less reliable
for structured outputs such as code, JSON, and molecular strings.
Proposed tokens can violate inter-token dependencies 
and degrade sequence-level consistency 
since multiple positions may be filled 
independently~\citep{KangGOLZZHHKCL2026,KimJJN2026}.
Furthermore, the nonsequential nature of DLM generation complicates syntax checking
during decoding.
Unlike autoregressive decoding, DLM decoding must reason about partially
filled sequences whose unresolved masks may appear between generated tokens.

Recent work has begun to develop constrained decoding for DLMs
motivated by these challenges.
\citet{SureshBUMS2025} studied regex constraints, 
while \citet{MuendlerDV2026} extended constrained DLM decoding 
to context-free grammars~(CFGs) through completion-based rejection sampling.
More recently, \citet{ZhangLLLJLL2026} proposed LAVE, which uses lookahead generation 
to verify candidate updates under CFG constraints.
These approaches demonstrate that CFG constraints can substantially improve 
structured output validity, but enforcing them introduces 
additional inference cost during DLM decoding.
This creates a central trade-off between structural correctness 
and the parallel efficiency of diffusion decoding.

We build on the rejection-sampling decoder of \citet{MuendlerDV2026}, 
which accepts a candidate update only when 
the remaining masked positions can still be completed into a sequence 
accepted by the target grammar. 
While effective, this procedure can be substantially 
slower than unconstrained decoding 
because each validity query requires repeated lexical and automaton processing.
Moreover, candidate updates are validated sequentially, 
limiting parallel token commitment.

To improve this trade-off, 
we propose \ours{}, a CFG-constrained decoding framework 
that reduces the major sources of overhead in this rejection-sampling pipeline, 
as illustrated in Figure~\ref{EPIC:main:fig:moti}.
\ours{} reuses lexical computations across similar partial outputs, 
performs DFA-free CFG validation directly on the lexed graph, 
and selects compatible candidate subsets for parallel commitment 
before final CFG verification. 
Together, these components reduce repeated lexing and validation costs 
while recovering parallel token commitment.

Our main contributions are as follows.

\begin{itemize}
    \item We identify three major overheads in rejection-sampling-based
    CFG-constrained diffusion decoding, namely repeated lexing, sequential candidate validation,
    and DFA construction.
    
    \item We propose \ours{}, a CFG-constrained decoding framework
    for DLMs that combines lexing memoization, relaxed compatible subset selection, 
    and DFA-free validation to reduce these costs 
    and recover parallel token commitment.

    \item We evaluate \ours{} against two CFG-constrained DLM decoders 
    across four models and three structured-generation benchmarks, 
    showing an improved trade-off between inference efficiency 
    and syntactic and functional correctness.
\end{itemize}

\section{Related Work}
\label{EPIC:main:sec:related_work}
We assume basic familiarity with regular languages, finite automata,
context-free grammars~(CFGs), lexing, and CFG parsing,
and refer to Appendix~\ref{EPIC:app:sec:preliminaries} for formal background.
We focus here on constrained decoding for diffusion language models~(DLMs)
and defer broader discussions of DLMs, AR constrained decoding,
and formal-language-based generation to
Appendix~\ref{EPIC:app:sec:additional_related_works}.

Constrained decoding for DLMs has recently progressed 
from regular-language to context-free constraints.
\citet{SureshBUMS2025} proposed DINGO for regular-language constraints,
while \citet{MuendlerDV2026} introduced CFG-constrained decoding
through completion-based rejection sampling.
Given a candidate update to a partially masked output,
the latter constructs an automaton representing its possible completions
and accepts the candidate when this completion language is compatible 
with the target CFG.
This formulation enables CFG constraints for nonsequential DLM generation,
but requires repeated lexing, automaton determinization and minimization,
and CFG compatibility checking.
Moreover, candidate updates are checked and committed sequentially,
reducing the parallelism available in diffusion decoding.

More recently, \citet{ZhangLLLJLL2026} proposed LAVE,
which approaches CFG-constrained DLM decoding 
through model-guided lookahead.
For each proposed token, LAVE samples several fillings of the remaining
masked positions from the distributions predicted by the DLM,
thereby constructing complete prefixes that can be checked 
by a standard grammar parser.
The candidate is accepted when at least one lookahead prefix is
grammatically extendable.
LAVE additionally caches verified prefixes 
to recover from repeatedly rejected proposals.
This provides an alternative to automata-based CFG completion checking,
but its proposal and verification procedure still commits candidate
tokens one at a time, with multiple lookahead prefixes considered
during each verification.

Our work focuses on reducing the computational overhead of CFG completion checking 
while recovering multi-token commitment.
Building on the completion-based formulation of
\citet{MuendlerDV2026}, \ours{} reuses lexical computations,
avoids DFA construction during exact CFG validation,
and uses a regular cover to identify candidate subsets that can be
verified and committed together.

\section{Problem Formulation}
\label{EPIC:main:sec:problem_formulation}

Let $G = (V, \Sigma_{\mathrm{lex}}, P, S)$
be a context-free grammar, where \(\Sigma_{\mathrm{lex}}\) is the
lexeme vocabulary, the set of grammar terminals.
The target constraint language is therefore $L(G) \subseteq \Sigma_{\mathrm{lex}}^*$.
This lexeme vocabulary should not be confused with either the byte-level
alphabet \(\Sigma_{\mathrm{byte}}\), used to represent raw strings, or the
tokenizer vocabulary \(\Gamma\) used by the diffusion language model.
A partial model output is represented as $x \in (\Gamma \cup \{\mask\})^n$,
where \(\mask\) denotes a mask.

Validity checking requires a lexical interface 
since the model operates over tokenizer-level sequences whereas the CFG is
defined over lexeme sequences.
We let $\Lex$ denote the lexical map that converts a token-level partial
output $x$ into a regular or graph representation of compatible lexeme
sequences.
We denote this representation by \(\GLang_x\), 
with $L(\GLang_x) \subseteq \Sigma_{\mathrm{lex}}^*.$
For notational convenience, we write
$C_x := L(\GLang_x)$
for the lexeme-sequence completion language induced by \(x\), and
$M(x) := \{ i \in [n] \mid x_i = \mask \}$
for the set of masked positions.

At each diffusion step, the model predicts distributions over tokens for masked positions.
Given a reveal budget \(k\), the decoder selects a subset of masked positions
$I = \{i_1,\ldots,i_k\} \subseteq M(x)$
and samples tokenizer tokens for these positions in parallel,
$t_i \sim p_\theta(\cdot \mid x), \quad i \in I.$
Equivalently, the model proposes updates
$\Delta = \{(i,t_i) \mid i \in I\}.$
Applying this update to \(x\) gives a new partial output defined
position-wise by
\[
(x \oplus \Delta)_j =
\begin{cases}
t_j & \text{if } (j,t_j)\in \Delta,\\
x_j & \text{otherwise}.
\end{cases}
\]

The central question in CFG-constrained DLM decoding is whether this
token-level update preserves the possibility of completing the remaining
masks into a lexeme sequence accepted by the grammar.
We say that a partial output \(x\) is completable with respect to \(G\) if
$L(G) \cap C_x \neq \emptyset .$
A proposed update \(\Delta\) is valid when
\[
\Valid_G(x,\Delta)
=
\mathbf{1}\left[
L(G)\cap C_{x\oplus\Delta}\neq \emptyset
\right].
\]

This definition gives a rule for constrained
diffusion decoding.
If \(\Valid_G(x,\Delta)=1\), the decoder may commit the proposed tokens.
Otherwise, the proposal must be rejected.
The completion-based baseline evaluates this condition by constructing an automaton for
\(C_{x\oplus\Delta}\) and checking its compatibility with the CFG.
Our goal is to compute the same validity predicate more efficiently in the
repeated and parallel setting induced by DLM decoding.

The multi-token nature of \(\Delta\) is crucial.
In unconstrained DLM decoding, all tokens in \(I\) can be committed
simultaneously after one model forward pass.
Under CFG constraints, however, a naive decoder must test the proposed
tokens sequentially, $(i_1,t_{i_1}), (i_2,t_{i_2}), \ldots, (i_k,t_{i_k})$,
because the validity of one token may depend on which other tokens have
already been accepted.
This sequential processing weakens the main advantage of DLMs.
The problem we address is therefore efficient validity checking for
multiple token updates, while preserving the CFG-completable output criterion 
over lexeme sequences.

\section{Proposed Method}
\label{EPIC:main:sec:proposed_method}
\begin{figure*}[th]
    \centering
    \includegraphics[width=\linewidth]{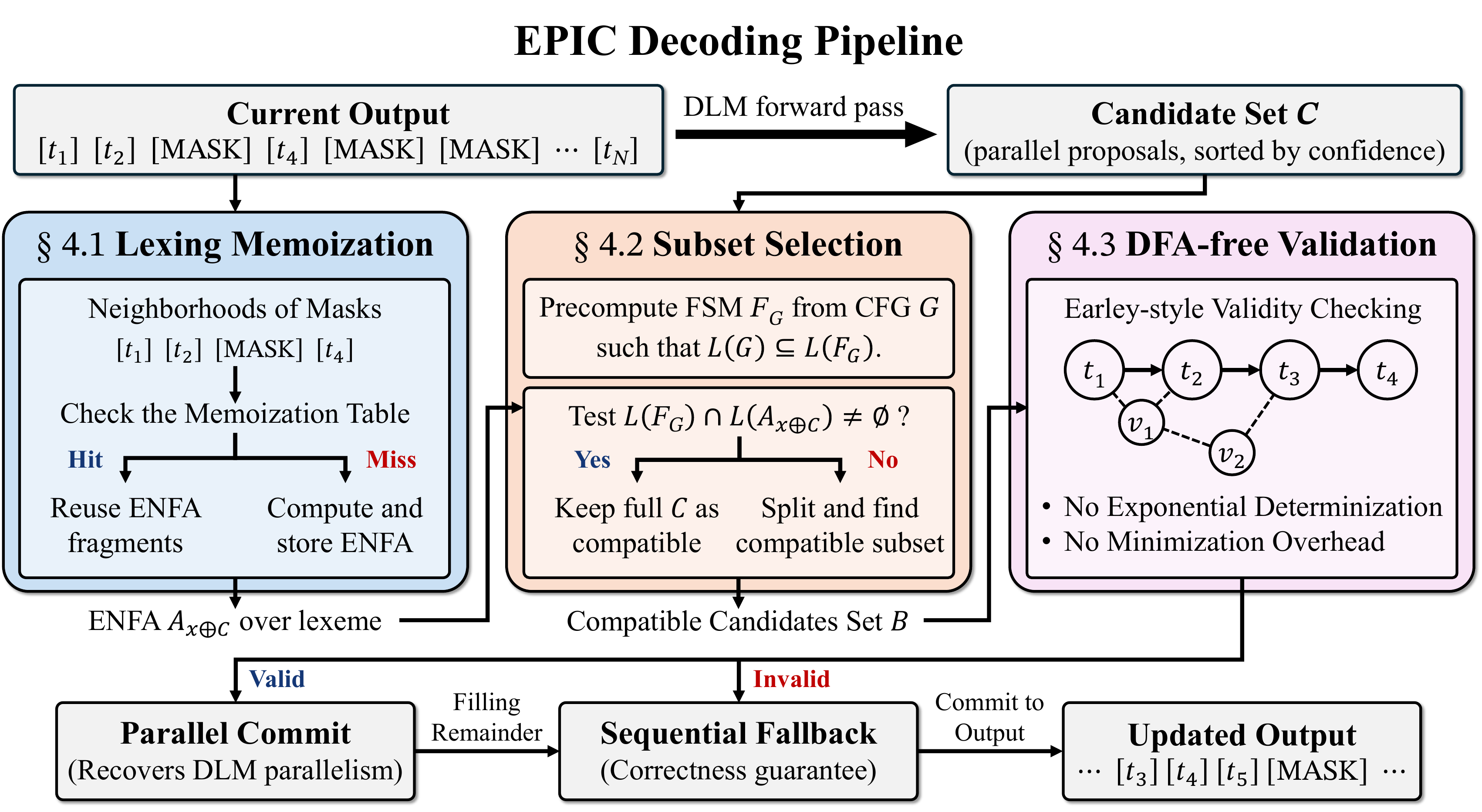}
    \caption{
    Overview of \ours{}.
    \ours{} combines lexing memoization, regular-cover batch selection, and DFA-free graph-parser validation to reduce the overhead of CFG-constrained decoding.
    }
    \label{EPIC:main:fig:overview}
\end{figure*}

We propose \textbf{\ours{}}, a decoding framework for \textbf{E}fficient and \textbf{P}arallel \textbf{I}nference
under CFG \textbf{C}on\-straints.
Our method consists of three main components, 
namely lexing memoization, 
relaxed compatible subset selection for parallel commit,
and DFA-free validation with Earley-style parsing.
Each component addresses a distinct source of overhead in prior work.
An overview of our framework is illustrated in Figure~\ref{EPIC:main:fig:overview}.

\subsection{Lexing Memoization}
The first component exploits locality in lexing.
Although masks introduce lexing ambiguity, 
each ambiguity is determined by a local text segment 
and its adjacent mask boundaries.
During diffusion decoding, an update modifies only a small part of the sequence,
while the remaining mask positions stay fixed.
As a result, consecutive validation queries often contain many unchanged 
local lexing contexts.
We therefore memoize possible prefix and suffix lexings 
for segments adjacent to masks.
When the same local context appears again, 
we reuse the stored lexing result during partial-language construction.
This avoids repeated lexing of unchanged regions and reduces the cost of
constructing the partial-output automaton or graph.

\subsection{Relaxed Compatible Subset Selection for Parallel Commit}
The second component is relaxed compatible subset selection, 
which addresses the sequential commit problem that significantly limits
the parallelism of DLMs.
When the number of denoising steps is small, a DLM reveals many tokens in a single step.
This is a key advantage of DLMs over AR language models, but it can also make structured generation more difficult because independently proposed tokens may be mutually incompatible.
Prior work performs sequential compatibility checking to ensure compatibility among independently sampled tokens.
As a result, when the number of denoising steps is small, existing methods incur a larger slowdown relative to unconstrained decoding.
The most direct solution would be to check the validity of multiple tokens simultaneously.
This approach, however, does not reveal which tokens are compatible when the check fails.
Finding the truly compatible subset would require a number of checks exponential in the number of tokens masked at the same time.

We avoid this cost by first selecting a subset that is compatible under a relaxed condition.
Our implementation builds a regular cover by flattening the production rules of the given CFG.
For a CFG~$G$, a regular cover is a finite automaton~$\FA$ that satisfies $L(G)\subseteq L(\FA)$.
Given a candidate batch $C$, we intersect this regular cover with the completion automaton induced by $x\oplus C$ and perform an emptiness check.
This relaxed check is cheaper than exact CFG validation and safely rejects candidate subsets 
that are incompatible even under the relaxed condition.
If the full batch is not cover-compatible, we recursively split the candidates by confidence and recover a smaller relaxed-compatible subset.
A relaxed-compatible subset is not committed immediately.
It is subsequently verified by the exact CFG completability checker,
described in the next subsection.
If exact verification rejects the selected subset, we recursively shrink it again 
and then fill the remaining reveal budget with the standard sequential constrained sampler.

\subsection{DFA-free Validation with Earley-style Parsing}
The third component is DFA-free validation.
Prior work represented a lexed partial output as an epsilon-NFA~(ENFA)
and then determinized and minimized it before CFG validation.
After this conversion, validation checked whether the automaton admitted
some path whose sequence of lexemes could be generated by the CFG.
Prior work implemented this search as dynamic programming over 
CFG nonterminals and automaton pairs, analogous to CYK-style
recognition~\citep{Younger1967,Kasami1966} on a finite automaton.
However, this pipeline incurs the cost of ENFA-to-DFA conversion at each
decoding step, and this bottom-up formulation provides limited
top-down guidance during the search.

In contrast, for exact validation, 
we avoid ENFA-to-DFA conversion by performing witness checking directly on the
ENFA using an Earley-style graph parser.
The parser combines bottom-up evidence from ENFA transitions 
with top-down predictions from the CFG.
Intuitively, the search meets in the middle.
The ENFA provides possible lexeme transitions, 
while the grammar restricts which partial derivations are relevant to the start symbol.
Epsilon transitions are traversed without advancing the grammar state,
while matching terminal-labeled transitions advance the corresponding parser item.
Although the worst-case complexity is not substantially different from 
that of a CYK-style implementation, 
the procedure avoids repeated determinization and minimization 
and introduces top-down grammar guidance into the witness search.
Thus, DFA-free validation preserves the completable output criterion
while removing the per-step overhead of ENFA-to-DFA conversion.
Detailed algorithms for the three main components of \ours{} are provided in
Appendix~\ref{EPIC:app:sec:detailed_algorithms}, and
Appendix~\ref{EPIC:app:sec:correctness} shows that these components preserve
the same completable-output criterion used by the baseline decoder.

\subsection{Failure recovery}
Following prior work, we use the same automatic recovery procedure 
when decoding reaches a fixed resampling budget.
This budget counts the number of rejected proposals over the entire decoding process.
Even when the budget is exhausted, 
the current partial output is known to be completable, 
since every accepted update has been verified with a witness.
Therefore, a valid witness completion always exists for the remaining masks.
Under the recovery procedure of prior work, 
model sampling is terminated early and the output is completed 
by extracting a valid completion from a witness.
This recovery step is not sampled from the model distribution, 
but provides a valid fallback 
when further rejection sampling is unlikely to succeed.

\section{Experimental Results and Analysis}
\label{EPIC:main:sec:experiments}

We evaluate \ours{} to show that 
it reduces the inference cost of CFG-constrained decoding for DLMs 
while maintaining syntactic and functional correctness.
Our experiments follow the evaluation protocol of prior work 
and cover three structured generation tasks: 
C++ code generation, JSON generation under instance-specific schemas, and SMILES generation.
We organize the evaluation around three questions.
First, we compare the relative inference time of \ours{} 
and the prior CFG-constrained baseline 
across different denoising-step settings, using unconstrained decoding as
the normalization reference.
Second, we analyze where runtime is spent 
and how each component of \ours{} reduces the runtime cost.
Third, we perform an ablation study over lexing memoization, DFA-free graph
validity checking, and relaxed compatible subset selection to isolate
their individual and combined effects.
We provide additional experimental details in Appendix~\ref{EPIC:app:sec:experimental_details} 
and additional results in Appendix~\ref{EPIC:app:sec:experimental_results}.

\subsection{Experimental Settings}
\label{EPIC:main:ssec:experimental_settings}

We evaluate the same four instruction-tuned DLMs as prior work, 
namely Dream, DreamCoder, LLaDA, and DiffuCoder.
The models include both general-purpose DLMs and code-oriented variants.
Following prior CFG-constrained decoding work, we use three structured-output
benchmarks, C++ code generation, JSON generation with instance-specific schemas,
and SMILES generation.
The C++ task is based on the HumanEval-X C++ split~\citep{ZhengXZDWXSWWLSYT2023},
while the JSON and SMILES tasks use the benchmarks from the prior evaluation
suite~\citep{MuendlerDV2026}.
We primarily compare \ours{} 
with the prior completion-based CFG-constrained baseline,
using unconstrained diffusion decoding as the reference 
for normalizing inference time.
\emph{Con.} denotes the prior CFG-constrained decoder, and \emph{\ours{}}
denotes our constrained decoding framework.
All constrained methods use the same task-specific grammars and lexical
specifications.

For the main comparison, we measure inference time 
across denoising steps in $\{16, 32, 64,\allowbreak 128, 256\}$ 
with a maximum generation length of $256$ tokens, and report the results
as relative inference time normalized by unconstrained decoding.
Smaller denoising step settings reveal more tokens per step and 
therefore place greater stress on the sequential validity checking bottleneck.
We use greedy decoding and repeat runtime measurements with
three seeds, $\{42,43,44\}$.
We additionally report syntactic and functional correctness after
automatic recovery to evaluate whether the efficiency gains of \ours{}
come at the cost of output correctness.

\begin{table*}[thb]
    \caption{
    Inference time relative to unconstrained decoding.
    For each method, we report the relative inference time normalized by
    unconstrained diffusion decoding, where $100\%$ indicates the same runtime as
    unconstrained decoding.
    Con. denotes the prior CFG-constrained decoder, and \ours{} denotes
    our method.
    The shaded column highlights \ours{}.
    }
    \label{EPIC:main:tab:main_results}
    \centering
    \small
    \begin{tabular}{crrrrrrrrr
    }
    \toprule[1.2pt]
        \multirow{2}{*}[-0.2em]{Model} & \multirow{2}{*}[-0.2em]{Steps} & 
        \multicolumn{2}{c}{C++} & \multicolumn{2}{c}{JSON} & \multicolumn{2}{c}{SMILES} & \multicolumn{2}{c}{Average}\\
        \cmidrule(lr){3-4}
        \cmidrule(lr){5-6}
        \cmidrule(lr){7-8}
        \cmidrule(lr){9-10}
        & & Con. & \ours{} & Con. & \ours{} & Con. & \ours{} & Con. & \ours{} \\
    \midrule[1.2pt]
        \multirow{5}{*}{\makecell[c]{Dream\\(7B)}} & 16 &
        143.86 & \CCG 98.44 & 176.32 & \CCG 109.36 & 102.09 & \CCG 103.43 & 140.76 & \CCG 103.74 \\
                                  & 32 &
        147.89 & \CCG 114.85 & 123.79 & \CCG 101.22 & 100.45 & \CCG 101.07 & 124.04 & \CCG 105.71 \\
                                  & 64 &
        114.41 & \CCG 108.12 & 108.19 & \CCG 101.36 & 100.67 & \CCG 101.18 & 107.76 & \CCG 103.55 \\
                                  & 128 &
        106.03 & \CCG 102.65 & 104.42 & \CCG 100.92 & 100.30 & \CCG 100.29 & 103.58 & \CCG 101.29 \\
                                  & 256 &
        102.12 & \CCG 102.92 & 102.39 & \CCG 100.82 & 98.96 & \CCG 99.21 & 101.16 & \CCG 100.98 \\
    \midrule
        \multirow{5}{*}{\makecell[c]{DreamCoder\\(7B)}} & 16 &
        388.60 & \CCG 127.34 & 178.21 & \CCG 89.19 & 101.14 & \CCG 103.14 & 222.65 & \CCG 106.56 \\
                                  & 32 &
        269.76 & \CCG 140.51 & 132.63 & \CCG 89.52 & 100.72 & \CCG 101.34 & 167.70 & \CCG 110.46 \\
                                  & 64 &
        182.52 & \CCG 128.04 & 118.02 & \CCG 96.74 & 100.41 & \CCG 100.65 & 133.65 & \CCG 108.48 \\
                                  & 128 &
        133.57 & \CCG 110.46 & 106.71 & \CCG 98.21 & 100.25 & \CCG 100.26 & 113.51 & \CCG 102.98 \\
                                  & 256 &
        117.23 & \CCG 106.50 & 101.26 & \CCG 98.84 & 100.03 & \CCG 100.01 & 106.17 & \CCG 101.78 \\
    \midrule
        \multirow{5}{*}{\makecell[c]{LLaDA\\(8B)}} & 16 &
        112.19 & \CCG 74.84 & 123.15 & \CCG 74.90 & 112.40 & \CCG 94.95 & 115.91 & \CCG 81.56 \\
                                  & 32 &
        95.49 & \CCG 64.07 & 111.99 & \CCG 77.51 & 102.77 & \CCG 92.22 & 103.42 & \CCG 77.93 \\
                                  & 64 &
        93.26 & \CCG 76.78 & 104.67 & \CCG 87.55 & 98.76 & \CCG 95.49 & 98.90 & \CCG 86.61 \\
                                  & 128 &
        100.52 & \CCG 93.27 & 102.42 & \CCG 95.21 & 98.17 & \CCG 96.03 & 100.37 & \CCG 94.84 \\
                                  & 256 &
        102.30 & \CCG 100.71 & 100.09 & \CCG 98.25 & 98.94 & \CCG 98.55 & 100.44 & \CCG 99.17 \\
    \midrule
        \multirow{5}{*}{\makecell[c]{DiffuCoder\\(7B)}} & 16 &
        187.50 & \CCG 108.07 & 165.58 & \CCG 108.72 & 103.93 & \CCG 104.96 & 152.34 & \CCG 107.25 \\
                                  & 32 &
        182.20 & \CCG 123.18 & 125.93 & \CCG 96.37 & 103.93 & \CCG 102.58 & 137.35 & \CCG 107.38 \\
                                  & 64 &
        138.87 & \CCG 102.33 & 111.23 & \CCG 97.45 & 103.36 & \CCG 101.01 & 117.82 & \CCG 100.26 \\
                                  & 128 &
        121.15 & \CCG 103.98 & 102.38 & \CCG 96.46 & 108.05 & \CCG 104.62 & 110.53 & \CCG 101.69 \\
                                  & 256 &
        118.54 & \CCG 107.80 & 95.17 & \CCG 92.97 & 103.17 & \CCG 102.10 & 105.63 & \CCG 100.96 \\
    \midrule
        \multicolumn{2}{r}{Average} & 147.90 & \CCG 104.74 & 119.73 & \CCG 95.58 & 101.93 & \CCG 100.15 & 123.18 & \CCG 100.16 \\
    \bottomrule[1.2pt]
    \end{tabular}
\end{table*}

\begin{table*}[t]
    \caption{
    Syntactic and functional correctness after automatic recovery.
    S. and F. denote syntactic and functional correctness, respectively,
    and Drm., Drm.C., L., and Diff.C denote Dream, DreamCoder, LLaDA, and DiffuCoder.
    The best result for each setting is shown in bold, and the shaded columns indicate \ours{}.
    All values are percentages.
    }
    \label{EPIC:main:tab:correctness}
    \centering
    \small
    \setlength{\tabcolsep}{2.3pt}
    \begin{tabular}{cr
        rrrrrr
        rrrrrr
        rrrrrr}
    \toprule[1.2pt]
        \multirow{3}{*}[-4pt]{Model} &
        \multirow{3}{*}[-4pt]{Step} &
        \multicolumn{6}{c}{C++} &
        \multicolumn{6}{c}{JSON} &
        \multicolumn{6}{c}{SMILES} \\
        \cmidrule(lr){3-8}
        \cmidrule(lr){9-14}
        \cmidrule(lr){15-20}

        & &
        \multicolumn{2}{c}{Uncon.} &
        \multicolumn{2}{c}{Con.} &
        \multicolumn{2}{c}{\ours{}} &
        \multicolumn{2}{c}{Uncon.} &
        \multicolumn{2}{c}{Con.} &
        \multicolumn{2}{c}{\ours{}} &
        \multicolumn{2}{c}{Uncon.} &
        \multicolumn{2}{c}{Con.} &
        \multicolumn{2}{c}{\ours{}} \\
        \cmidrule(lr){3-4}
        \cmidrule(lr){5-6}
        \cmidrule(lr){7-8}
        \cmidrule(lr){9-10}
        \cmidrule(lr){11-12}
        \cmidrule(lr){13-14}
        \cmidrule(lr){15-16}
        \cmidrule(lr){17-18}
        \cmidrule(lr){19-20}

        & &
        S. & F. & S. & F. & S. & F. &
        S. & F. & S. & F. & S. & F. &
        S. & F. & S. & F. & S. & F. \\
    \midrule[1.2pt]

    \multirow{5}{*}{Drm.}
    & 16
    & 35.4 & 10.4 & \textbf{100} & \textbf{11.6} & \CCG \textbf{100} & \CCG \textbf{11.6}
    & 32.4 & 13.2 & \textbf{100} & \textbf{23.7} & \CCG \textbf{100} & \CCG \textbf{23.7}
    & 61.7 & \textbf{0.6} & \textbf{100} & \textbf{0.6} & \CCG \textbf{100} & \CCG \textbf{0.6} \\
    & 32
    & 67.7 & 15.2 & \textbf{100} & \textbf{16.5} & \CCG \textbf{100} & \CCG \textbf{16.5}
    & 77.6 & 43.4 & \textbf{100} & \textbf{46.0} & \CCG \textbf{100} & \CCG \textbf{46.0}
    & 70.7 & \textbf{3.0} & \textbf{100} & \textbf{3.0} & \CCG \textbf{100} & \CCG \textbf{3.0} \\
    & 64
    & 80.5 & 18.1 & \textbf{99.6} & 18.5 & \CCG \textbf{99.6} & \CCG \textbf{18.9}
    & 92.3 & 53.7 & \textbf{100} & \textbf{54.8} & \CCG \textbf{100} & \CCG \textbf{54.8}
    & 72.5 & 3.0 & \textbf{100} & \textbf{3.6} & \CCG \textbf{100} & \CCG \textbf{3.6} \\
    & 128
    & 89.0 & 13.8 & \textbf{100} & \textbf{14.6} & \CCG 99.8 & \CCG 14.4
    & 90.4 & 51.5 & \textbf{100} & \textbf{53.3} & \CCG \textbf{100} & \CCG \textbf{53.3}
    & 73.1 & 3.6 & \textbf{100} & \textbf{4.2} & \CCG \textbf{100} & \CCG \textbf{4.2} \\
    & 256
    & 82.9 & 12.8 & \textbf{100} & 13.4 & \CCG \textbf{100} & \CCG \textbf{14.0}
    & 91.2 & 51.8 & \textbf{100} & \textbf{53.7} & \CCG \textbf{100} & \CCG \textbf{53.7}
    & 72.5 & 3.0 & \textbf{100} & \textbf{4.2} & \CCG \textbf{100} & \CCG \textbf{4.2} \\

    \midrule

    \multirow{5}{*}{Drm.C.}
    & 16
    & 17.7 & 6.7 & \textbf{99.8} & 8.5 & \CCG \textbf{99.8} & \CCG \textbf{9.1}
    & 79.0 & 41.9 & \textbf{100} & \textbf{44.1} & \CCG \textbf{100} & \CCG 44.0
    & 78.4 & \textbf{4.2} & \textbf{100} & \textbf{4.2} & \CCG \textbf{100} & \CCG \textbf{4.2} \\
    & 32
    & 54.3 & 19.5 & \textbf{99.4} & 20.7 & \CCG \textbf{99.4} & \CCG \textbf{21.3}
    & 87.5 & 52.2 & \textbf{100} & \textbf{53.3} & \CCG \textbf{100} & \CCG \textbf{53.3}
    & 79.0 & 3.6 & \textbf{100} & \textbf{4.2} & \CCG \textbf{100} & \CCG \textbf{4.2} \\
    & 64
    & 72.6 & 26.8 & 99.2 & \textbf{27.2} & \CCG \textbf{100} & \CCG 26.6
    & 90.8 & \textbf{53.7} & \textbf{100} & 53.3 & \CCG \textbf{100} & \CCG 53.3
    & 74.9 & \textbf{2.4} & \textbf{100} & \textbf{2.4} & \CCG \textbf{100} & \CCG \textbf{2.4} \\
    & 128
    & 73.8 & 28.0 & \textbf{100} & \textbf{28.7} & \CCG \textbf{100} & \CCG \textbf{28.7}
    & 93.0 & 55.9 & \textbf{100} & \textbf{56.6} & \CCG \textbf{100} & \CCG \textbf{56.6}
    & 74.3 & \textbf{2.4} & \textbf{100} & \textbf{2.4} & \CCG \textbf{100} & \CCG \textbf{2.4} \\
    & 256
    & 84.1 & \textbf{28.0} & \textbf{100} & 27.4 & \CCG \textbf{100} & \CCG 27.4
    & 94.1 & 56.3 & \textbf{100} & \textbf{56.6} & \CCG \textbf{100} & \CCG \textbf{56.6}
    & 74.3 & \textbf{2.4} & \textbf{100} & \textbf{2.4} & \CCG \textbf{100} & \CCG \textbf{2.4} \\

    \midrule

    \multirow{5}{*}{L.}
    & 16
    & 5.5 & 0.6 & 99.6 & 4.1 & \CCG \textbf{100} & \CCG \textbf{4.5}
    & 57.4 & 39.0 & \textbf{100} & \textbf{44.1} & \CCG \textbf{100} & \CCG \textbf{44.1}
    & 41.3 & 1.2 & \textbf{100} & \textbf{1.8} & \CCG \textbf{100} & \CCG \textbf{1.8} \\
    & 32
    & 9.1 & 3.0 & 99.2 & 5.3 & \CCG \textbf{99.4} & \CCG \textbf{6.3}
    & 75.4 & 44.1 & \textbf{100} & \textbf{50.0} & \CCG \textbf{100} & \CCG \textbf{50.0}
    & 60.5 & 0.6 & \textbf{100} & \textbf{1.2} & \CCG \textbf{100} & \CCG \textbf{1.2} \\
    & 64
    & 42.7 & 10.4 & 99.8 & \textbf{11.0} & \CCG \textbf{100} & \CCG \textbf{11.0}
    & 79.8 & 44.5 & \textbf{100} & \textbf{50.0} & \CCG \textbf{100} & \CCG \textbf{50.0}
    & 65.9 & 1.2 & \textbf{100} & \textbf{1.8} & \CCG \textbf{100} & \CCG \textbf{1.8} \\
    & 128
    & 82.9 & \textbf{15.9} & \textbf{100} & \textbf{15.9} & \CCG \textbf{100} & \CCG \textbf{15.9}
    & 87.9 & 48.2 & \textbf{100} & \textbf{52.9} & \CCG \textbf{100} & \CCG \textbf{52.9}
    & 65.9 & \textbf{1.8} & \textbf{100} & \textbf{1.8} & \CCG \textbf{100} & \CCG \textbf{1.8} \\
    & 256
    & 96.3 & \textbf{25.6} & \textbf{100} & 25.0 & \CCG \textbf{100} & \CCG 25.0
    & 91.2 & 49.6 & \textbf{100} & \textbf{52.9} & \CCG \textbf{100} & \CCG \textbf{52.9}
    & 65.3 & \textbf{2.4} & \textbf{100} & \textbf{2.4} & \CCG \textbf{100} & \CCG \textbf{2.4} \\

    \midrule

    \multirow{5}{*}{Diff.C}
    & 16
    & 25.0 & 12.8 & \textbf{99.8} & 14.0 & \CCG \textbf{99.8} & \CCG \textbf{14.6}
    & 76.8 & 42.3 & \textbf{100} & \textbf{44.1} & \CCG \textbf{100} & \CCG \textbf{44.1}
    & 70.7 & 0.0 & \textbf{100} & \textbf{0.6} & \CCG \textbf{100} & \CCG \textbf{0.6} \\
    & 32
    & 36.0 & 15.2 & \textbf{99.8} & \textbf{18.3} & \CCG \textbf{99.8} & \CCG 17.7
    & 87.9 & 47.8 & \textbf{100} & \textbf{49.6} & \CCG \textbf{100} & \CCG 49.3
    & 71.9 & 0.0 & \textbf{100} & 0.0 & \CCG \textbf{100} & \CCG 0.0 \\
    & 64
    & 53.7 & 17.7 & 99.8 & \textbf{18.3} & \CCG \textbf{100} & \CCG \textbf{18.3}
    & 84.6 & 45.2 & \textbf{100} & \textbf{47.4} & \CCG \textbf{100} & \CCG \textbf{47.4}
    & 62.3 & \textbf{0.6} & \textbf{100} & \textbf{0.6} & \CCG \textbf{100} & \CCG \textbf{0.6} \\
    & 128
    & 60.4 & 25.6 & \textbf{100} & \textbf{29.9} & \CCG 99.8 & \CCG 29.7
    & 77.9 & 42.3 & \textbf{100} & \textbf{43.8} & \CCG \textbf{100} & \CCG \textbf{43.8}
    & 49.7 & \textbf{1.8} & \textbf{100} & \textbf{1.8} & \CCG \textbf{100} & \CCG \textbf{1.8} \\
    & 256
    & 93.9 & 43.9 & \textbf{100} & 45.5 & \CCG \textbf{100} & \CCG \textbf{45.7}
    & 77.9 & 41.9 & \textbf{100} & \textbf{43.4} & \CCG \textbf{100} & \CCG \textbf{43.4}
    & 47.9 & \textbf{2.4} & \textbf{100} & \textbf{2.4} & \CCG \textbf{100} & \CCG \textbf{2.4} \\

    \bottomrule[1.2pt]
    \end{tabular}
\end{table*}

\subsection{Main Results}
Tables~\ref{EPIC:main:tab:main_results} and
\ref{EPIC:main:tab:correctness} summarize the main results.
Table~\ref{EPIC:main:tab:main_results} reports relative inference time
for four models and three datasets, normalized by unconstrained decoding,
while Table~\ref{EPIC:main:tab:correctness} reports syntactic and
functional correctness after automatic recovery.
Thus, a value of $100$ indicates the same runtime as unconstrained decoding,
while smaller values indicate faster decoding.
Across C++ and JSON, \ours{} outperforms the prior
CFG-constrained baseline in $39$ of the $40$ settings.
In several cases, \ours{} achieves relative inference time below $100$.
This does not imply negative constraint checking cost.
Rather, constrained decoding can terminate early 
once a complete valid output and EOS have been formed, 
whereas the unconstrained diffusion sampler executes the full configured schedule.
This effect is most visible on JSON and LLaDA.
The improvement of \ours{} is especially pronounced on C++, where the grammar
is more complex than those used for JSON and SMILES and requires more production
rules and automaton states.

On SMILES, the relative-time difference between the baseline and \ours{} is
generally small, except for LLaDA.
This is because the fixed CFG and lexical specification used for SMILES 
are much smaller than those for the other tasks, 
so the baseline already stays close to unconstrained decoding.
As a result, there is less room for our optimizations 
to reduce per-check cost.
The LLaDA case is different because of its block-wise decoding schedule.
LLaDA divides the canvas into blocks 
and fills a fixed number of masks within each block, 
whereas Dream, DreamCoder, and DiffuCoder determine 
the number and positions of unmasked tokens 
dynamically from a global timestep schedule.
This makes the candidate sets of LLaDA more localized and stable, which can make
compatible subset selection more effective.
Thus, even though graph parsing and caching provide limited gains on SMILES,
LLaDA can still benefit from a reduction in sequential exact checks.
We provide further analysis in
Section~\ref{EPIC:main:ssec:further_analysis}.

The benefit of \ours{} is larger when the number of denoising steps is small.
This matches our motivation, 
since smaller step counts require the decoder to unmask more tokens per step 
and therefore amplify the sequential validation bottleneck.
The effect is particularly clear for DreamCoder.
For example, on DreamCoder with C++ at $16$ denoising steps, 
the prior CFG-constrained baseline requires $388.60\%$ 
of the unconstrained decoding time,
whereas \ours{} requires only $127.34\%$.
Equivalently, the overhead beyond unconstrained decoding is reduced from
$288.60\%$ to $27.34\%$.

\begin{table*}[t]
    \caption{
    Runtime breakdown of the baseline and \ours{}.
    We report the time spent in lexing, ENFA-to-DFA conversion, and
    validation.
    Validation includes both CFG checks and regular cover checks.
    For graph parser variants, DFA time comes from regular cover selection.
    }
    \label{EPIC:main:tab:runtime_analysis}
    \centering
    \small
    \setlength{\tabcolsep}{3.0pt}
    \begin{tabular}{ccrrrrrrrrr}
    \toprule[1.2pt]
        \multirow{2}{*}[-2pt]{Model} &
        \multirow{2}{*}[-2pt]{Method} &
        \multicolumn{3}{c}{C++} &
        \multicolumn{3}{c}{JSON} &
        \multicolumn{3}{c}{SMILES} \\
        \cmidrule(lr){3-5}
        \cmidrule(lr){6-8}
        \cmidrule(lr){9-11}
        & &
        Lex.~(s) & DFA~(s) & Val.~(s) &
        Lex.~(s) & DFA~(s) & Val.~(s) &
        Lex.~(s) & DFA~(s) & Val.~(s) \\
    \midrule[1.2pt]

    \multirow{2}{*}{Dream}
        & Con. &
        50.8 & 47.2 & 49.8 &
        24.4 & 27.1 & 2.0 &
        0.3 & 1.0 & 0.7 \\
        & \ours{} &
        \CCG 12.0 & \CCG 11.8 & \CCG 17.1 &
        \CCG 5.9 & \CCG 5.6 & \CCG 2.1 &
        \CCG 0.1 & \CCG 0.2 & \CCG 0.5 \\
    \midrule

    \multirow{2}{*}{DreamCoder}
        & Con. &
        96.9 & 216.8 & 179.8 &
        17.3 & 31.0 & 1.7 &
        0.3 & 0.3 & 0.1 \\
        & \ours{} &
        \CCG 33.4 & \CCG 68.5 & \CCG 36.7 &
        \CCG 4.3 & \CCG 6.4 & \CCG 1.8 &
        \CCG 0.1 & \CCG 0.0 & \CCG 0.4 \\
    \midrule

    \multirow{2}{*}{LLaDA}
        & Con. &
        47.1 & 20.0 & 56.6 &
        21.7 & 7.5 & 1.0 &
        2.6 & 3.0 & 33.1 \\
        & \ours{} &
        \CCG 21.6 & \CCG 3.7 & \CCG 11.6 &
        \CCG 4.2 & \CCG 1.0 & \CCG 0.9 &
        \CCG 1.5 & \CCG 0.4 & \CCG 7.1 \\
    \midrule

    \multirow{2}{*}{DiffuCoder}
        & Con. &
        83.4 & 102.8 & 146.2 &
        26.2 & 40.4 & 2.5 &
        1.2 & 1.8 & 18.4 \\
        & \ours{} &
        \CCG 18.3 & \CCG 18.1 & \CCG 25.0 &
        \CCG 5.3 & \CCG 8.2 & \CCG 2.6 &
        \CCG 0.2 & \CCG 0.2 & \CCG 1.6 \\
    \bottomrule[1.2pt]
    \end{tabular}
\end{table*}

Table~\ref{EPIC:main:tab:correctness} shows that the substantial runtime
reductions of \ours{} do not come at the cost of syntactic correctness.
After automatic recovery, both constrained methods achieve nearly perfect
syntactic correctness across all models, datasets, and denoising-step settings.
The few residual failures occur when witness extraction
does not finish within the configured timeout.
\ours{} closely matches the prior CFG-constrained baseline, with differences
limited to at most $0.2$ percentage points when \ours{} is lower.
This indicates that recovering multi-token commitment does not meaningfully
weaken the structural correctness provided by CFG-constrained decoding.

Functional correctness shows the same overall pattern.
Despite changing the token commitment order and committing multiple compatible
tokens together, \ours{} remains closely comparable to the sequential
CFG-constrained baseline across all settings.
The largest decrease relative to the baseline is only $0.6$ percentage points,
while \ours{} also improves functional correctness in several settings.
Together with the runtime results in Table~\ref{EPIC:main:tab:main_results}, 
these results show that \ours{} improves decoding efficiency 
while largely preserving both syntactic and functional correctness.

\begin{table*}[t]
    \caption{
    Component analysis of \ours{}.
    Memoization reports reused lexing results and hit rate, 
    while parallelism reports the average selected subset size 
    and regular cover selection success rate.
    }
    \label{EPIC:main:tab:components_analysis}
    \centering
    \small
    \setlength{\tabcolsep}{3.5pt}
    \begin{tabular}{crrrrrrrrrrrr}
    \toprule[1.2pt]
        \multirow{3}{*}[-4pt]{Model} &
        \multicolumn{4}{c}{C++} &
        \multicolumn{4}{c}{JSON} &
        \multicolumn{4}{c}{SMILES} \\
        \cmidrule(lr){2-5}
        \cmidrule(lr){6-9}
        \cmidrule(lr){10-13}
        &
        \multicolumn{2}{c}{Memoization} &
        \multicolumn{2}{c}{Parallelism} &
        \multicolumn{2}{c}{Memoization} &
        \multicolumn{2}{c}{Parallelism} &
        \multicolumn{2}{c}{Memoization} &
        \multicolumn{2}{c}{Parallelism} \\
        \cmidrule(lr){2-3}
        \cmidrule(lr){4-5}
        \cmidrule(lr){6-7}
        \cmidrule(lr){8-9}
        \cmidrule(lr){10-11}
        \cmidrule(lr){12-13}
        &
        \#~Hit & Rate & \#~Tok. & Rate &
        \#~Hit & Rate & \#~Tok. & Rate &
        \#~Hit & Rate & \#~Tok. & Rate \\
    \midrule[1.2pt]

    Dream &
    141,128 & 67.2 & 3.4 & 97.4 &
    757,257 & 88.9 & 2.7 & 93.9 &
    5,116 & 43.3 & 2.2 & 97.1 \\
    
    DreamCoder &
    429,345 & 75.2 & 3.7 & 93.5 &
    1,178,326 & 95.3 & 4.9 & 95.0 &
    4,590 & 44.5 & 1.9 & 91.6 \\
    
    LLaDA &
    62,347 & 53.1 & 5.1 & 94.5 &
    269,782 & 83.5 & 7.0 & 97.1 &
    13,884 & 46.5 & 3.6 & 87.8 \\
    
    DiffuCoder &
    300,019 & 72.3 & 3.7 & 90.9 &
    970,458 & 90.9 & 3.9 & 95.2 &
    31,442 & 65.0 & 2.9 & 92.1 \\

    \bottomrule[1.2pt]
    \end{tabular}
\end{table*}

\subsection{Further Analysis}
\label{EPIC:main:ssec:further_analysis}
We further analyze \ours{} to verify that 
its empirical gains come from the intended sources of efficiency improvement.
All results in this subsection are measured at $32$ denoising steps.
We first examine the runtime breakdown 
in Table~\ref{EPIC:main:tab:runtime_analysis}.
The table compares the baseline and \ours{} in terms of
lexing time, DFA-construction time, and validation time.
\ours{} reduces overhead in almost all components.
Lexing memoization yields roughly a $2\times$ to $4\times$ reduction 
in lexing time, with the largest gains appearing on C++ and JSON, 
where lexing accounts for a substantial fraction of the baseline overhead.
The DFA-construction time also decreases substantially.
Although \ours{} does not construct a DFA 
for exact validation of each partial output, 
it still constructs automata during relaxed compatible subset selection.
This cost remains small 
because regular-cover checks are used only for parallel commit, 
while the sequential fallback runs without DFA construction.
Validation time is also reduced, especially on C++ and LLaDA.
C++ benefits from DFA-free validation 
because its grammar is more complex than the grammars used for JSON and SMILES.
LLaDA benefits more strongly from validation reduction 
since its fixed-number unmasking policy makes parallel commit more stable
than in the other models.

We next report component-level statistics of \ours{} 
in Table~\ref{EPIC:main:tab:components_analysis}.
Memoization achieves hit rates above $50\%$ in most settings, 
and C++ and JSON show particularly large numbers of reused lexing results.
These high hit rates indicate 
that diffusion decoding repeatedly visits similar local lexing contexts 
and that the memoization table captures meaningful reuse.
The parallelism statistics show that
regular-cover selection succeeds in about $94\%$ of selection attempts.
They also show that \ours{} selects more than two tokens 
for parallel commitment on average across all models and datasets.
The effect is most pronounced for LLaDA.
Unlike the other models, LLaDA reveals a fixed number of tokens 
in each decoding block, which provides more stable candidates
for relaxed compatible subset selection.
On JSON, LLaDA commits an average of $7.0$ tokens in parallel, 
demonstrating that \ours{} can recover substantial diffusion parallelism 
under CFG constraints.

\begin{table*}[t]
    \caption{
        Ablation results for \ours{}.
        Starting from the baseline~(Con.),
        we separately add $\circled{1}$ lexing memoization,
        $\circled{2}$ DFA-free validation with Earley-style parsing,
        and $\circled{3}$ relaxed compatible subset selection for parallel commit.
        \ours{} enables all three components.
    }
    \label{EPIC:main:tab:ablation_results}
    \centering
    \small
    \setlength{\tabcolsep}{2.8pt}
    \begin{tabular}{lrrrrrrrrrrrrrrr}
    \toprule[1.2pt]
        \multirow{2}{*}{Model} &
        \multicolumn{5}{c}{C++} &
        \multicolumn{5}{c}{JSON} &
        \multicolumn{5}{c}{SMILES} \\
        \cmidrule(lr){2-6}
        \cmidrule(lr){7-11}
        \cmidrule(lr){12-16}
        &
        Con. & $\circled{1}$ & $\circled{2}$ & $\circled{3}$ & \ours{} &
        Con. & $\circled{1}$ & $\circled{2}$ & $\circled{3}$ & \ours{} &
        Con. & $\circled{1}$ & $\circled{2}$ & $\circled{3}$ & \ours{} \\
    \midrule[1.2pt]

    Dream &
        900 & 843 & 863 & 900 & \CCG \textbf{700} &
        1203 & 1095 & 1237 & 1162 & \CCG \textbf{983} &
        \textbf{516} & 541 & 540 & 545 & \CCG 519 \\

    DreamCoder &
        1582 & 1385 & 1261 & 1378 & \CCG \textbf{824} &
        1089 & 939 & 1109 & 986 & \CCG \textbf{735} &
        \textbf{523} & 540 & 540 & 543 & \CCG \textbf{523} \\

    LLaDA &
        791 & 772 & 738 & 601 & \CCG \textbf{531} &
        1554 & 1450 & 1619 & 1179 & \CCG \textbf{1076} &
        636 & 655 & 628 & 609 & \CCG \textbf{570} \\

    DiffuCoder &
        1278 & 1152 & 1085 & 1119 & \CCG \textbf{862} &
        1314 & 1172 & 1307 & 1195 & \CCG \textbf{1005} &
        584 & 598 & 582 & 601 & \CCG \textbf{576} \\

    \bottomrule[1.2pt]
    \end{tabular}
\end{table*}

\subsection{Ablation Results}
\label{EPIC:main:ssec:ablation_results}
Table~\ref{EPIC:main:tab:ablation_results} shows the ablation results at
$32$ denoising steps.
We evaluate each component of \ours{} by adding it separately to the baseline.
The results further confirm the observations 
from the runtime breakdown analysis.
The components of \ours{} provide distinct sources of speedup, 
and their effects become clearer in the ablation study.
The gains are especially pronounced on C++.
All three components improve or match inference time for every model.
This consistent improvement indicates that \ours{} is particularly effective
when the constraint language is complex enough to make repeated lexing,
automaton construction, and exact CFG validation costly.
On JSON, DFA-free validation alone sometimes increases runtime.
This is likely because JSON grammars are simpler than the C++ grammar, 
so the baseline intersection-based validation already incurs 
relatively small overhead.
In this setting, the additional graph-parser overhead can outweigh 
the reduction from avoiding DFA construction.
On SMILES, the gains are limited because the baseline constraint
overhead is already small.
The full \ours{} configuration generally achieves the best inference time by
combining the complementary benefits of the three components.
Pairwise combinations further show incremental improvements over
single component variants in most settings, as reported in the full ablation
results in Appendix~\ref{EPIC:app:ssec:ablation_results}.

\section{Conclusions}

We presented \ours{}, an efficient CFG-constrained decoding framework 
for diffusion language models.
\ours{} reduces the main overheads of prior constrained diffusion decoding
through lexing memoization, relaxed compatible subset selection for parallel
commit, and DFA-free validation with Earley-style graph parsing.
These components preserve the completable output criterion 
while reducing repeated lexing, avoiding per-step DFA construction for exact validation, 
and recovering parallel token commitment.
Experiments on C++, JSON, and SMILES with four DLMs show that 
\ours{} substantially reduces inference time 
while maintaining correctness comparable to the baseline.
At $16$ denoising steps, where parallelism matters most, 
\ours{} achieves a $3.1\times$ speedup over the prior CFG-constrained decoder 
on DreamCoder for C++ and reduces the relative inference time from
$388.6\%$ to $127.3\%$ of unconstrained decoding.
Component analysis further shows memoization hit rates above $50\%$ in most settings 
and an average selected subset size above two tokens across all models and datasets.
Future work could further reduce overhead through incremental graph parsing,
tighter modeling of finite token budgets, 
and extensions beyond CFGs to lightweight context-sensitive constraints 
such as type or schema-dependent semantic checks.

\subsection*{AI use statement}

In this work, we used generative AI tools to assist with the implementation
of experimental code.
We have not used generative AI tools for other tasks requiring disclosure,
and the rest of the required-disclosure tasks are not applicable to this work.
Additionally, we used generative AI tools to edit the manuscript for
readability and language polishing.
We have reviewed all AI-assisted work.
AI-assisted experimental code was manually reviewed and tested for correctness,
and all AI-assisted manuscript edits were reviewed by the authors.
We take responsibility for the final content of this work,
including text, claims or artifacts produced with the aid of generative AI.

\bibliography{iclr2027_conference}
\bibliographystyle{iclr2027_conference}

\appendix

\section{Preliminaries}

\label{EPIC:app:sec:preliminaries}
This section reviews the formal language theory 
and compiler theory background used throughout the paper. 
We focus on the concepts necessary to define CFG-constrained decoding, 
analyze the computational bottlenecks, 
and analyze the correctness of our proposed procedures.

\subsection{Regular Languages and Finite Automata}

A regular language is a set of strings that can be recognized by a finite automaton, 
or described 
by a regular expression~\citep{Sipser1997, HopcroftU1979}. 
A DFA is defined as a tuple $(Q,\Sigma,\delta,q_0,F)$, 
where $Q$ is a finite set of states, $\Sigma$ is a finite alphabet, 
$\delta: Q \times \Sigma \rightarrow Q$ is a transition function, 
$q_0 \in Q$ is the initial state, 
and $F \subseteq Q$ is the set of accepting states. 
An NFA generalizes a DFA 
by allowing multiple outgoing transitions for the same input symbol 
and may also include $\varepsilon$-transitions. 
Every NFA can be converted to an equivalent DFA via subset construction, 
although the resulting DFA may contain exponentially many states 
in the worst-case~\citep{Sipser1997, HopcroftU1979}.

As in prior work, we use NFAs to represent the partial outputs with masked regions.
In prior CFG-constrained decoding pipelines, this NFA is determinized and
then minimized before CFG compatibility checking.
For a DFA with state set \(Q\) over alphabet \(\Sigma\), Hopcroft's algorithm
runs in \(O(|\Sigma||Q| \log |Q|)\) time in the worst-case~\citep{Hopcroft1971}.
Since the automaton is reconstructed from many partial outputs during decoding, 
repeated determinization and minimization become a non-negligible source of overhead.

\subsection{Context-Free Grammars}

A CFG is defined as a tuple $G=(V,\Sigma,P,S)$, 
where $V$ is a finite set of nonterminals, 
$\Sigma$ is a finite set of terminals, 
$P$ is a set of productions of the form $A \to \alpha$ 
with $A \in V$ and $\alpha \in (V \cup \Sigma)^*$, 
and $S \in V$ is the start symbol. 
The language generated by $G$ is denoted by $L(G)$. 
Context-free languages~(CFLs) are more expressive than regular languages, and 
can express recursive structures such as balanced parentheses, 
nested expressions, and programming languages.

Prior work uses a closure property that the intersection of a CFL 
and a regular language is context-free~\citep{Sipser1997}. 
This property is used to formulate validity checking as 
an emptiness test over the intersection between the target CFL
and the possible completions of the partial output. 
In such a construction, the intersection grammar has nonterminals 
indexed by CFG nonterminals and pairs of automaton states, 
which introduces a cubic dependence on the number of automaton states.
This leads to a validity-checking cost of $O\left(|P||Q|^3 + |P||\Sigma||Q|^2\right)$.

Our relaxed check follows the classical idea of finite-state or regular
approximation of CFG
~\citep{PereiraW1991,Nederhof2000,MohriN2001}.
We view the CFG as a recursive transition network, 
where each nonterminal corresponds to a subautomaton 
with an entry state and an exit state.
We then flatten this network into a finite automaton 
by replacing each recursive call to a nonterminal with a transition to 
that nonterminal's entry state, 
and by connecting its exit state to the possible continuation states.
After flattening, the automaton no longer records 
which call site entered a nonterminal.
Consequently, the automaton may accept additional lexeme sequences, but every
CFG derivation is still preserved as a valid path.
The resulting automaton is therefore a regular cover \(\FA\) satisfying
\[
L(G) \subseteq L(\FA).
\]

\subsection{Lexers and Parsers}

Practical parsers rarely operate on raw character sequences directly. 
A typical compiler front end first applies lexing, 
which maps a character-level string into a sequence of tokens or lexemes, 
and then applies parsing to verify 
whether the token sequence conforms to a grammar. 
Additional context-sensitive checks, 
such as type checking or name resolution,
are often performed after parsing.

This separation is important for constrained decoding 
since language models generate tokens at the token level, 
whereas CFGs are usually defined over lexemes. 
Partial outputs with masked regions make lexing ambiguous, 
since a generated token may belong to a lexeme on its left, 
a lexeme on its right, or a lexeme spanning a masked region.

Earley parsing is a general CFG parsing algorithm 
that combines top-down prediction 
with bottom-up completion~\citep{Earley1970}.
Unlike purely bottom-up methods such as CYK, 
Earley-style parsing can use top-down grammar predictions 
to restrict which partial derivations are considered during parsing.
Our DFA-free validity checker adopts this principle 
for witness checking over graph-structured partial outputs.

\section{Additional Related Work}
\label{EPIC:app:sec:additional_related_works}

This section provides additional related work 
that complements Section~\ref{EPIC:main:sec:related_work}. 
We discuss diffusion language models, 
constrained decoding for autoregressive language models, 
and constrained decoding methods for diffusion language models.

\subsection{Diffusion Language Models}

Diffusion language models~(DLMs) provide an alternative 
to autoregressive text generation~\citep{AustinJHTB2021, LouME2024, SahooASGMCRK2024, NieZYZOHZLWL2025}. 
Instead of generating tokens from left to right, 
DLMs iteratively refine a partially observed sequence, 
often starting from a sequence of masked tokens. 
At each denoising step, the model predicts tokens 
for one or more masked positions conditioned 
on the current partially filled sequence.

Masked diffusion language models~(MDLMs) instantiate this idea 
using a discrete masking process~\citep{SahooASGMCRK2024, NieZYZOHZLWL2025}. 
During generation, a model repeatedly replaces masked positions 
with sampled tokens until the sequence is fully specified. 
This generation process enables nonsequential 
and potentially parallel token prediction, 
since multiple masked positions can be updated
within the same denoising step.
The degree of parallelism is typically controlled 
by the number of denoising steps 
or the number of tokens committed per step. 
Fewer steps allow more tokens to be generated per forward pass, 
improving speed but often increasing the risk of inconsistent or low-quality outputs~\citep{KimJJN2026}.

This parallel generation makes DLMs attractive for efficient decoding.
At the same time, it creates new challenges for structured generation. 
Since multiple tokens may be sampled independently before being committed, 
their joint compatibility with syntactic constraints is not guaranteed. 
This issue motivates constrained decoding methods 
that can preserve the parallelism of DLMs 
while enforcing formal language constraints.

\subsection{Constrained Decoding for AR Models}

Constrained decoding aims to ensure that model outputs 
conform to a predefined set of constraints. 
In many applications, this set is specified by a formal language, 
such as a regular expression, a JSON schema, or a context-free grammar. 
Constrained decoding has been widely studied 
for autoregressive language models, 
where generation proceeds from left to right 
and the validity of each next token can be checked against the prefix~\citep{ScholakSB2021, GengJPW2023, ParkZD2025}.

For regular constraints, 
valid next tokens can be computed 
using finite automata~\citep{KooLH2024, WillardL2023}. 
For context-free constraints, 
the decoder must account for nested and recursive structures, 
which requires parsing-based or grammar-based methods~\citep{GengJPW2023,BeurerFV2024,UgareSKMS2025}. 
Existing approaches differ in whether they mask invalid tokens before sampling, 
reject invalid samples after sampling, or combine both strategies~\citep{WillardL2023,BeurerFV2024,UgareSKMS2025}. 
These methods are effective for autoregressive generation 
because the prefix grows monotonically from left to right.
The DLM setting breaks this assumption. 
A partial output may contain multiple masks, 
and updates can occur at arbitrary positions. 
As a result, prefix-based constrained decoding algorithms cannot be directly applied. 
This motivates validity checking procedures that reason about 
whether a partially filled sequence can still be completed into a valid string.

\subsection{Constrained Decoding for DLMs}

Several recent works have begun to study constrained decoding 
for diffusion language models.
\citet{SureshBUMS2025} proposed DINGO, a constrained inference method 
for DLMs under regular-language constraints.
Their method showed that formal constraints can be integrated 
into non-autoregressive diffusion decoding.
However, regular constraints cannot capture many structured output settings
that require nested or recursive structure, such as nested JSON objects,
arithmetic expressions, and programming-language syntax.

The closest prior work to ours is \citet{MuendlerDV2026}, 
which introduced CFG-constrained decoding for diffusion LLMs. 
Their framework formulates validity checking 
as an infilling-based completion
and reduces it to checking the non-emptiness of the intersection 
between a CFG and a regular language 
representing possible completions of the partial output. 
This approach enables CFG constraints for diffusion language models 
and multi-region infilling, 
but it requires repeated lexing, automaton construction, DFA minimization,
and intersection-based emptiness checking during decoding. 
Their paper explicitly constructs an NFA for the partial output, 
converts it to a DFA, minimizes it, 
and then performs CFG intersection-based checking.

\section{Detailed Algorithms}
\label{EPIC:app:sec:detailed_algorithms}

We provide implementation-level pseudocode for \ours{}.
These algorithms clarify how \ours{} preserves 
the completable-output criterion of prior CFG-constrained decoding 
while changing how this criterion is evaluated during diffusion decoding.
Algorithm~\ref{EPIC:app:alg:epic_decoding} presents the main decoding loop,
including proposal generation, relaxed subset selection, exact verification,
and token commitment.
Algorithm~\ref{EPIC:app:alg:lexing_memoization} describes how lexical
representations are stored and reused across partial outputs.
Algorithm~\ref{EPIC:app:alg:graph_parser} specifies the Earley-style graph
parser used for DFA-free validity checking.
Algorithm~\ref{EPIC:app:alg:relaxed_subset_selection} gives 
the compatible subset selection for multiple-token proposals 
under the relaxed regular cover.

\begin{algorithm*}[t!]
\caption{\ours{} Decoding under CFG Constraints}
\label{EPIC:app:alg:epic_decoding}
\begin{algorithmic}[1]
\Require Prompt $p$, diffusion LM $M$, CFG $G$, lexical map $\Lex$, answer length $L$, steps $T$
\Ensure Output tokens $x$

\State{Compile $\Lex$ and initialize the lexing memoization table.}
\State{Construct a regular cover $\FA$ by flattening the production rules of $G$.}
\State{Initialize $x$ with $p$ followed by $L$ copies of \mask{}.}

\ForAll{diffusion blocks}
    \For{$t=1,\ldots,T$}
        \Lcomment{Parallel proposal from the diffusion model}
        \State{Run $M$ on $x$ and score proposals for the currently eligible masked positions.}
        \State{Compute the reveal budget $n_t$ for this denoising step.}
        \State{Form $C$ from up to $n_t$ highest-confidence eligible non-EOS proposals.}

        \Lcomment{Select a relaxed-compatible batch and verify it exactly before committing}
        \State{$B \gets \Call{RelaxedSubsetSelect}{x,C,\FA,G,\Lex}$.}
        \State{Commit all candidates in $B$ to $x$.}

        \Lcomment{Fall back to sequential rejection sampling 
        if verification fails or commits remain}
        \For{$j=1,\ldots,n_t-|B|$}
            \State{Select the next highest-confidence candidate $c$ not already committed.}
            \While{$\Call{DFAFreeValidate}{x\oplus c,G,\Lex}$ is false}
                \State{Reject $c$, suppress its logit, and select the next candidate.}
            \EndWhile
            \State{Commit the accepted candidate $c$ to $x$.}
        \EndFor
    \EndFor
\EndFor

\State \Return $x$
\end{algorithmic}
\end{algorithm*}

\subsection{Overall Decoding Procedure}

Algorithm~\ref{EPIC:app:alg:epic_decoding} summarizes 
the overall decoding procedure of \ours{}.
At each denoising step, the model scores the currently masked positions 
and proposes candidate tokens for multiple positions.
The role of the constrained decoder is to decide 
which of these updates can be committed 
without losing the possibility of completing the sequence 
into a valid string in $L(G)$.
The main difference from the baseline lies in 
how this validity decision is organized.
The baseline effectively checks candidate tokens one at a time 
using exact CFG validation.
This sequentializes the decoding process 
and weakens the main advantage of diffusion language models.
In contrast, \ours{} first calls \Call{RelaxedSubsetSelect}{} to search 
for a compatible subset that may be committed together.
This subset selection step is intentionally conservative. 
It may return a smaller or empty batch, 
but any nonempty batch must still pass exact verification before commitment.

\begin{algorithm*}[th]
\caption{Lexing Memoization for Partial Outputs}
\label{EPIC:app:alg:lexing_memoization}
\begin{algorithmic}[1]
\Require Partial output $x$, lexical map $\Lex$, lexing cache $\mathcal{C}_{\mathrm{lex}}$
\Ensure $\varepsilon$-NFA $\GLang_x$ representing possible lexeme sequences

\State{Merge consecutive generated tokens in $x$ into text spans separated by masked regions.}
\State{Initialize an empty $\varepsilon$-NFA $\GLang_x$.}

\ForAll{fixed text spans $w$ in order}
    \If{$(w,\mathrm{boundary}) \in \mathcal{C}_{\mathrm{lex}}$}
        \State{Retrieve the cached lexing result for $w$.}
    \Else
        \State{Compute the lexing result for $w$ using $\Lex$.}
        \State{Store the computed result in $\mathcal{C}_{\mathrm{lex}}$.}
    \EndIf
    \State{Add span-level transitions to $\GLang_x$ using the retrieved or computed lexing result.}
\EndFor

\State{Complete the construction by adding transitions for masked gaps and lexemes spanning them.}
\State \Return $\GLang_x$
\end{algorithmic}
\end{algorithm*}

For candidates that are not committed through the batch filter, 
the algorithm falls back to the exact rejection loop.
This fallback is important for two reasons.
First, it preserves the behavior of rejection-based constrained decoding 
when parallel commitment is not possible.
Second, it separates efficiency from correctness.
Relaxed compatible subset selection is used only to propose a subset 
that is likely to pass exact validation, 
thereby reducing the number of sequential checks.
The final accept-or-reject decision is still made by
\Call{DFAFreeValidate}{}.
Thus, every token committed by
Algorithm~\ref{EPIC:app:alg:epic_decoding} is still justified 
by the completable-output condition.

\subsection{Lexing Memoization}

Algorithm~\ref{EPIC:app:alg:lexing_memoization} describes 
the lexing memoization procedure.
Since the CFG is defined over lexemes 
while the diffusion model operates over a tokenizer vocabulary, 
validity checking first requires an $\varepsilon$-NFA representation of lexeme sequences
compatible with the current masked string.
The automaton construction follows 
the partial-output lexing procedure, 
which handles fixed text spans, masked gaps, and lexemes 
that may span masked gaps.

\ours{} applies memoization to the lexing of text spans.
For each span, the procedure first checks 
whether the same span has already been lexed 
under the same boundary condition.
If the result is cached, 
it is reused during the construction of $\GLang_x$.
If the entry is missing, the span is lexed using the lexical map $\Lex$,
and the result is inserted into $\mathcal{C}_{\mathrm{lex}}$.

The output of the procedure is an $\varepsilon$-NFA $\GLang_x$.
Paths in this automaton correspond to lexeme sequences 
that are compatible with the current partial output.
Lexing memoization does not approximate or change this represented language.
It only changes the construction procedure 
by avoiding repeated computation for local contexts 
that have already been processed.
Thus, lexing memoization reduces the cost of constructing $\GLang_x$ 
without changing the completable-output check.

\begin{algorithm*}[th]
\caption{DFA-Free Validation with Earley-style Parsing}
\label{EPIC:app:alg:graph_parser}
\begin{algorithmic}[1]
\Require Partial output $x$, CFG $G=(V,\Sigma,P,S)$, lexical map $\Lex$
\Ensure Whether $x$ can be completed into a string in $L(G)$

\State{$\GLang_x \gets \Call{MemoizedLexing}{x,\Lex,\mathcal{C}_{\mathrm{lex}}}$.}
\State{Initialize a worklist with Earley-style items at the start nodes of $\GLang_x$.}
\State{Initialize an empty visited set for parser items.}

\While{the worklist is not empty}
    \State{Pop an item from the worklist.}
    \If{the item has already been visited}
        \State{Continue to the next item.}
    \EndIf
    \State{Mark the item as visited.}
    \Lcomment{Traverse $\varepsilon$-transitions without advancing the grammar state}
    \ForAll{$\varepsilon$-edges from the item's current graph node to $q'$}
        \State{Add the same parser item at graph node $q'$ to the worklist.}
    \EndFor

    \If{the item expects a nonterminal}
        \Lcomment{Prediction}
        \State{Add predicted items for productions of the expected nonterminal to the worklist.}
    \ElsIf{the item expects a terminal}
        \Lcomment{Scanning over graph edges}
        \State{Follow matching terminal-labeled edges in $\GLang_x$ and add advanced items to the worklist.}
    \Else
        \Lcomment{Completion}
        \State{Advance items that were waiting for the completed nonterminal at the same graph node.}
        \State{Add the advanced items to the worklist.}
    \EndIf

    \If{an accepting item for the start symbol reaches an accepting node of $\GLang_x$}
        \State \Return true
    \EndIf
\EndWhile

\State \Return false
\end{algorithmic}
\end{algorithm*}

\subsection{DFA-Free Validation with Earley-style Parsing}

Algorithm~\ref{EPIC:app:alg:graph_parser} describes the DFA-free validity checker 
used by \ours{}.
After lexing memoization constructs the graph $\GLang_x$, 
the goal is to determine whether the graph contains at least one path 
whose terminal sequence belongs to $L(G)$.
Equivalently, the procedure checks 
whether the partial output still has a completion satisfying the target CFG.

The baseline performs this check through an automata-theoretic pipeline.
It first converts the graph representation of the partial output into a DFA, 
often applies minimization, and then checks the non-emptiness of the intersection 
between the DFA language and the CFG language.
This approach is correct, but it introduces substantial repeated overhead.
In particular, determinization and minimization are performed many times 
during decoding.

\ours{} avoids repeated DFA construction 
by running the witness search directly on $\GLang_x$.
The validator generalizes Earley-style parsing from a single input string
to an $\varepsilon$-NFA whose paths represent possible lexeme sequences.
Each parser item records both a CFG progress state and graph-node information, 
so that it represents a partial derivation over a path segment of $\GLang_x$.
Prediction introduces productions that can satisfy the next expected nonterminal,
$\varepsilon$-transitions move an item between graph nodes without advancing
its grammar position,
scanning advances items along compatible terminal-labeled edges, 
and completion connects a recognized constituent to items 
that were waiting for it at the corresponding graph node.
The algorithm accepts once it finds a completed start-symbol item 
that connects a start node of $\GLang_x$ to an accepting node.

This DFA-free formulation preserves the exact witness-search problem.
The algorithm does not ask whether one fixed string is generated by $G$.
Instead, it asks whether any path in the graph is generated by $G$.
This is precisely the same condition captured 
by the CFG and DFA intersection test.
This change should be understood as a practical optimization 
rather than an improvement 
in the worst-case asymptotic complexity of validity checking.
It reduces overhead by searching for the witness directly on $\GLang_x$
and avoiding construction of an equivalent minimal DFA 
at every validation.

\begin{algorithm*}[th]
\caption{Relaxed Compatible Subset Selection for Parallel Commit}
\label{EPIC:app:alg:relaxed_subset_selection}
\begin{algorithmic}[1]
\Require Partial output $x$, candidates $C$, regular cover $\FA$, CFG $G$, lexical map $\Lex$
\Ensure Exactly verified candidate subset $B$ with $|B|\ge 2$, or $\emptyset$

\Function{RelaxedSelect}{$x,C$}
    \If{$|C|\le1$}
        \State \Return $\emptyset$
    \EndIf

    \State{$x_C \gets x\oplus C$}
    \State{$\GLang_{x_C} \gets \Call{MemoizedLexing}{x_C,\Lex,\mathcal{C}_{\mathrm{lex}}}$}

    \If{$L(\FA)\cap L(\GLang_{x_C})\neq\emptyset$}
        \State \Return $C$ \Rcomment{Keep the full subset as relaxed-compatible}
    \EndIf

    \State{Split $C$ into high-confidence halves $C_L$ and $C_R$.}
    \State{$B_L \gets \Call{RelaxedSelect}{x,C_L}$}
    \State{$B_R \gets \Call{RelaxedSelect}{x\oplus B_L,C_R}$}
    \State \Return $B_L \cup B_R$
\EndFunction

\Function{ExactShrink}{$x,B$}
    \If{$|B|\le1$}
        \State \Return $\emptyset$
    \EndIf
    \If{$\Call{DFAFreeValidate}{x\oplus B,G,\Lex}$ is true}
        \State \Return $B$
    \EndIf
    \State{Split $B$ into high-confidence halves $B_L$ and $B_R$.}
    \State{$B'_L \gets \Call{ExactShrink}{x,B_L}$}
    \State{$B'_R \gets \Call{RelaxedSelect}{x\oplus B'_L,B_R}$}
    \State{$B''_R \gets \Call{ExactShrink}{x\oplus B'_L,B'_R}$}
    \State \Return $B'_L \cup B''_R$
\EndFunction

\State{Sort $C$ by descending confidence.}
\State{$B \gets \Call{RelaxedSelect}{x,C}$}
\State \Return $\Call{ExactShrink}{x,B}$
\end{algorithmic}
\end{algorithm*}

\subsection{Relaxed Compatible Subset Selection for Parallel Commit}

Algorithm~\ref{EPIC:app:alg:relaxed_subset_selection} describes 
the candidate subset selection procedure 
used to recover parallel token commitment.
For a candidate set $C$, we write $x\oplus C$ for the partial output 
obtained by applying all candidate updates in $C$ to $x$.
Checking and committing every proposed token sequentially weakens the
parallelism of DLM decoding.
Although a joint CFG check over all candidates is possible, 
a failed check does not identify which candidates caused the failure.
Finding the largest compatible subset by exhaustive search would require
a number of checks exponential in the number of proposed tokens.

\ours{} avoids this cost by selecting a subset 
under a cheaper relaxed condition before exact verification.
The relaxed condition is defined by a regular cover $\FA$ of the CFG language.
The automaton $\FA$ can be viewed as a finite-state relaxation of 
a recursive transition network representation of $G$. 
It ignores the return sites of recursive calls, which may introduce
additional paths but preserves every CFG derivation.
Therefore, $L(G)\subseteq L(\FA)$.
This inclusion makes rejection under the regular cover sound.
If a candidate batch is rejected by $\FA$, 
then applying the whole batch cannot yield a CFG-completable partial output.
If a candidate batch is accepted by $\FA$, 
it may still be invalid under the original CFG, 
so exact verification is required.

The recursive structure of Algorithm~\ref{EPIC:app:alg:relaxed_subset_selection} uses 
the regular cover as a cheap group test.
The algorithm first tests the whole candidate set under the relaxed condition.
When the whole set fails the cover check, 
it splits the candidates by confidence 
and searches for smaller high-confidence subsets 
that pass the relaxed check.
This allows the decoder to find a jointly plausible subset 
without enumerating all subsets.
The procedure is biased toward high-confidence candidates 
because those are more likely to be selected 
by the unconstrained diffusion decoder as well.
The final exact check is essential 
because the regular cover may introduce false positives.
Before a nonempty subset is committed, 
the algorithm verifies the partial output obtained 
after applying the subset using the exact CFG validity checker.
If a relaxed-compatible subset fails exact checking, 
the algorithm recursively shrinks it using the exact CFG validator.
Thus, before any nonempty subset is committed, 
the partial output obtained after applying the subset has been verified 
by the exact CFG validity checker.
The selector returns only nontrivial batches. 
Singleton candidates are left to the standard sequential rejection sampler, 
which verifies them with the exact CFG checker.
After committing the verified subset, 
the decoder uses the standard sequential rejection loop 
to make the remaining commits.
This fallback preserves correctness 
while allowing \ours{} to recover parallel commitment 
whenever a sufficiently large compatible subset is found.

\section{Correctness Analysis of Proposed Methods}
\label{EPIC:app:sec:correctness}

This section analyzes why \ours{} preserves the same completable-output
criterion as the baseline decoder.
The analysis consists of three parts.
First, lexing memoization reuses local lexing results 
without changing the partial-output language.
Second, DFA-free validation checks the same non-emptiness condition 
as the CFL and the regular language intersection test, 
but searches directly over the partial-output graph.
Third, the finite automaton used in relaxed subset selection 
is a regular cover of the CFG language.

\subsection{Lexing Memoization}

\emph{Lexing memoization preserves the partial-output language.}
For any partial output $x$, Algorithm~\ref{EPIC:app:alg:lexing_memoization}
constructs an NFA $\GLang_x$ that recognizes the same set of lexeme
sequences as the non-memoized lexing procedure.

The reason is that memoization changes only 
how local lexing alternatives are obtained.
For each local context, the algorithm 
either retrieves a previously computed result from the cache 
or computes the result using the same lexical map~$\Lex$.
Since cache entries are keyed 
by the corresponding local prefix and suffix configuration,
a cache hit refers to the same lexing subproblem.
The final graph is assembled 
from the same local alternatives 
and the same $\epsilon$-connections across masked regions.
Thus, memoization reduces construction cost 
but does not approximate or change the represented lexeme sequence language.

\subsection{DFA-Free Validation}

\emph{DFA-free validation checks exact CFG completability over the
partial-output graph.}
For a partial-output graph $\GLang_x$ and CFG $G$,
Algorithm~\ref{EPIC:app:alg:graph_parser} searches for a path in
$\GLang_x$ whose label sequence belongs to $L(G)$.
Thus, it targets the same condition
\[
L(G)\cap L(\GLang_x)\neq \emptyset .
\]

The validator can be viewed 
as an Earley-style recognizer generalized 
from string positions to graph nodes.
Prediction introduces productions 
that may satisfy the next expected nonterminal,
$\varepsilon$-transitions move items 
between graph nodes without consuming a grammar terminal,
scanning advances items along matching terminal-labeled edges,
and completion connects recognized constituents to items 
waiting for them.
An accepting item therefore corresponds to a CFG derivation 
along some path of $\GLang_x$.
Conversely, any such derivation can be explored 
by the same prediction, scanning, and completion operations.
The algorithm therefore changes the search strategy, 
not the validity criterion.

\subsection{Regular Cover}

\emph{The flattened automaton is a regular cover of the CFG language.}
Let $\FA$ be the finite automaton obtained by flattening 
the recursive transition network induced by a CFG $G$.
Then the automaton over-approximates the CFG language:
\[
L(G)\subseteq L(\FA).
\]

This construction follows the standard RTN-based regular superset
approximation of CFGs~\citep{Nederhof2000}.
A CFG can be viewed as a recursive transition network 
by introducing entry and exit states for each nonterminal 
and rule-position states for each production.
Flattening joins these components into a single finite automaton 
and replaces the recursive call-return mechanism with ordinary transitions.
This removes return-site matching and can therefore introduce additional paths, 
but it does not remove any path corresponding to a valid CFG derivation.

The inclusion follows by a structural induction over CFG derivation trees.
For each subtree rooted at a nonterminal $A$, 
the corresponding RTN contains a path 
from the entry state of $A$ to the exit state of $A$ 
whose labels are exactly the yield of that subtree.
For a production $A\to X_1\cdots X_m$, 
terminal symbols are matched by the corresponding labeled rule-position transitions,
and nonterminal children are matched 
by the induction hypothesis together with the flattened call 
and return transitions.
Since flattening keeps all such transitions 
and only forgets which call site a return must match, 
the yield of the whole derivation tree labels an accepting path in $\FA$.
Thus every string generated by $G$ is accepted by $\FA$.

\section{Experimental Details}
\label{EPIC:app:sec:experimental_details}

We follow the benchmark suite, constraints, and prompting protocol of prior
CFG-constrained decoding work for DLMs.
All main experiments use a maximum generation length of $256$ tokens and
temperature $0.0$ with greedy decoding.
For the main comparison, we evaluate denoising steps in $\{16,32,64,128,256\}$ 
and repeat runtime measurements with seeds $\{42,43,44\}$, 
although greedy decoding makes the generated outputs seed independent.
For each seed, we normalize the runtime of each constrained method
by the unconstrained runtime under the same model, task, and decoding-step
setting, and then average the relative runtimes across seeds.
For the ablation study, we fix the denoising step count to $32$ and evaluate
all subsets of the three acceleration components of \ours{}.

\subsection{Implementation and Decoding Settings}
\label{EPIC:app:ssec:implementation_details}

We implement \ours{} on top of the public implementation of prior
CFG-constrained decoding for diffusion language models~\citep{MuendlerDV2026}.
The evaluation pipeline, dataset wrappers, model wrappers, 
and profiling scripts are written in Python, 
while the formal-language backend uses Rust components for finite automata, 
CFG operations, regular-expression compilation, automaton construction, 
and intersection-related routines.

We compare three decoding modes.
Unconstrained decoding disables grammar constraints.
The baseline enables CFG constraints 
but disables the three acceleration components.
\ours{} enables lexing memoization, DFA-free validation, 
and relaxed compatible subset selection.
For the main comparison, we evaluate all three modes across the three tasks,
four models, five denoising-step settings, and three runtime seeds.
Smaller step counts reveal more tokens per denoising step 
and therefore place more pressure on the sequential checking bottleneck.
The ablation study is performed at $32$ denoising steps 
and varies only which of the three acceleration components are enabled.
For each run, we record generated outputs, total wall-clock time, 
and component-level profiling statistics used in the runtime analysis.

Wall-clock time is measured around each generation call, 
excluding model and tokenizer loading, common grammar and lexer preprocessing, 
and downstream correctness evaluation.
Regular-cover construction is performed lazily within the decoding path and
is therefore included in \ours{}'s runtime.
For C++ and SMILES, which use fixed task-level grammars, 
the constructed cover is reused across instances.
For JSON, each instance uses a schema-specific grammar, and the corresponding
cover is constructed separately for each instance.

\subsection{Datasets, Prompts, and Metrics}
\label{EPIC:app:ssec:datasets_prompts_metrics}

We evaluate three structured-generation tasks: C++ code generation, JSON
generation under instance-specific schemas, and SMILES generation.
Each prompt follows the prior constrained-diffusion implementation 
and contains a system instruction, a user input, 
and an assistant-side target prefix.
The assistant-side prefix places the model inside the target format 
before generation begins, 
so that the generated continuation aligns with the grammar 
used by the constrained decoder.

For C++, we use the C++ split of HumanEval-X~\citep{ZhengXZDWXSWWLSYT2023}.
Each instance provides a programming problem, a target function declaration,
and test cases.
We use the C++ CFG and lexical specification from the prior implementation.
Syntactic correctness is measured by CFG-based syntax checking, 
and functional correctness is measured by compiling the generated program 
and running the provided tests.

For JSON, we use the JSON benchmark from the prior CFG-constrained diffusion
evaluation suite~\citep{MuendlerDV2026}, which extends
JSON-Mode-Eval~\citep{Nousresearch2024}.
Each instance contains a natural-language input 
and an instance-specific JSON schema.
The schema is converted into an instance-specific CFG 
and lexical map using the prior implementation.
Syntactic correctness is measured by schema-CFG acceptance, 
and functional correctness is measured 
by comparing the normalized generated JSON object with the reference output.

For SMILES~\citep{Weininger1988}, we use the benchmark introduced 
by prior CFG-constrained diffusion work~\citep{MuendlerDV2026}.
Each instance asks the model to generate a SMILES string 
from a molecule description.
We use the SMILES CFG and lexical specification from the prior implementation.
Unlike C++ and JSON, this setting does not allow arbitrary whitespace 
between lexed tokens.
Syntactic correctness is measured by validity under the SMILES grammar 
and lexer, and functional correctness is measured by molecular equivalence 
to the reference SMILES string after canonicalization.

\subsection{Models}
\label{EPIC:app:ssec:model_details}

We evaluate four instruction-tuned diffusion language models following prior
CFG-constrained decoding work:
Dream~\citep{YeXZGWJLK2025},
DreamCoder~\citep{XieYZGDWZGJLK2025},
LLaDA~\citep{NieZYZOHZLWL2025}, and
DiffuCoder~\citep{GongZZGJKZ2026}.
Dream and LLaDA are general-purpose DLMs, while DreamCoder and DiffuCoder are
code-oriented DLMs.
The evaluated checkpoints are
Dream\footnote{\texttt{Dream-org/Dream-v0-Instruct-7B}.},
DreamCoder\footnote{\texttt{Dream-org/Dream-Coder-v0-Instruct-7B}.},
LLaDA\footnote{\texttt{GSAI-ML/LLaDA-8B-Instruct}.}, and
DiffuCoder\footnote{\texttt{apple/DiffuCoder-7B-Instruct}.}.
All models are evaluated with their corresponding tokenizer and wrapper.
Dream, DreamCoder, and DiffuCoder use the Dream-style generation interface.
LLaDA uses a separate wrapper 
because its chat template and generation interface differ.
These wrapper-level differences affect prompt formatting 
but not the task instructions, constraints, or evaluation protocol.

\subsection{Experimental Environments}
\label{EPIC:app:ssec:experimental_environments}

All experiments were conducted on a workstation with an
NVIDIA RTX A6000 GPU and an AMD Ryzen Threadripper 3960X CPU.
The software environment consisted of Rocky Linux~8.10, Python~3.11.15,
Rust~1.93.1, and PyTorch~2.12.0 built against CUDA~13.0.
For model and dataset handling, we used
\texttt{transformers}~4.52.2,
\texttt{datasets}~3.6.0, and
\texttt{accelerate}~1.13.0.
Rust extension modules were built with \texttt{maturin}~1.13.3.
For task-specific evaluation, we used \texttt{partialsmiles}~2.0 and
\texttt{rdkit}~2026.3.2.

\begin{table*}[thb]
    \caption{
        Full inference time results.
        We report wall-clock inference time in seconds for unconstrained decoding
        (Uncon.), the prior CFG-constrained baseline (Con.), and \ours{}.
        These timings correspond to the relative inference times reported in
        Table~\ref{EPIC:main:tab:main_results}.
        The results cover all models, tasks, and denoising-step settings used in
        the main comparison.
        The shaded column highlights \ours{}.
    }
    \label{EPIC:app:tab:full_inference_time_results}
    \centering
    \small
    \setlength{\tabcolsep}{5.5pt}
    \begin{tabular}{crrrrrrrrrr
    }
    \toprule[1.2pt]
        \multirow{2}{*}[-0.2em]{Model} & \multirow{2}{*}[-0.2em]{Steps} & 
        \multicolumn{3}{c}{C++} & \multicolumn{3}{c}{JSON} & \multicolumn{3}{c}{SMILES}\\
        \cmidrule(lr){3-5}
        \cmidrule(lr){6-8}
        \cmidrule(lr){9-11}
        & & Uncon. & Con. & \CCG \ours{} & Uncon. & Con. & \CCG \ours{} & Uncon. & Con. & \CCG \ours{} \\
    \midrule[1.2pt]
        \multirow{5}{*}{Dream~7B} & 16 &
        571 & 656 & \CCG 450 & 549 & 969 & \CCG 601 & 232 & 237 & \CCG 240 \\
                                  & 32 &
        609 & 900 & \CCG 700 & 972 & 1203 & \CCG 983 & 514 & 516 & \CCG 519 \\
                                  & 64 &
        1089 & 1245 & \CCG 1177 & 1822 & 1971 & \CCG 1847 & 1015 & 1022 & \CCG 1028 \\
                                  & 128 &
        2044 & 2167 & \CCG 2098 & 3476 & 3630 & \CCG 3508 & 1998 & 2004 & \CCG 2003 \\
                                  & 256 &
        3958 & 4042 & \CCG 4071 & 6812 & 6975 & \CCG 6867 & 4043 & 4001 & \CCG 4011 \\
    \midrule
        \multirow{5}{*}{DreamCoder~7B} & 16 &
        594 & 1876 & \CCG 609 & 517 & 921 & \CCG 461 & 240 & 242 & \CCG 247 \\
                                  & 32 &
        586 & 1582 & \CCG 824 & 821 & 1089 & \CCG 735 & 519 & 523 & \CCG 526 \\
                                  & 64 &
        1068 & 1949 & \CCG 1368 & 1530 & 1806 & \CCG 1480 & 1012 & 1016 & \CCG 1019 \\
                                  & 128 &
        2024 & 2703 & \CCG 2236 & 2833 & 3023 & \CCG 2782 & 2000 & 2005 & \CCG 2005 \\
                                  & 256 &
        3806 & 4462 & \CCG 4054 & 5479 & 5548 & \CCG 5415 & 3983 & 3984 & \CCG 3984 \\
    \midrule
        \multirow{5}{*}{LLaDA~8B} & 16 &
        520 & 578 & \CCG 386 & 923 & 1137 & \CCG 690 & 361 & 405 & \CCG 343 \\
                                  & 32 &
        828 & 791 & \CCG 531 & 1388 & 1554 & \CCG 1076 & 619 & 636 & \CCG 570 \\
                                  & 64 &
        1336 & 1246 & \CCG 1026 & 2332 & 2441 & \CCG 2042 & 1156 & 1141 & \CCG 1104 \\
                                  & 128 &
        2382 & 2394 & \CCG 2221 & 4233 & 4335 & \CCG 4030 & 2238 & 2197 & \CCG 2149 \\
                                  & 256 &
        4359 & 4459 & \CCG 4389 & 8023 & 8030 & \CCG 7882 & 4384 & 4337 & \CCG 4320 \\
    \midrule
        \multirow{5}{*}{DiffuCoder~7B} & 16 &
        628 & 1114 & \CCG 657 & 605 & 1002 & \CCG 657 & 349 & 362 & \CCG 366 \\
                                  & 32 &
        701 & 1278 & \CCG 862 & 1044 & 1314 & \CCG 1005 & 562 & 584 & \CCG 576 \\
                                  & 64 &
        1137 & 1579 & \CCG 1163 & 1902 & 2116 & \CCG 1854 & 1010 & 1043 & \CCG 1020 \\
                                  & 128 &
        1897 & 2298 & \CCG 1973 & 3601 & 3687 & \CCG 3474 & 1952 & 2109 & \CCG 2042 \\
                                  & 256 &
        3108 & 3684 & \CCG 3351 & 7046 & 6706 & \CCG 6551 & 3736 & 3854 & \CCG 3814 \\
    \bottomrule[1.2pt]
    \end{tabular}
\end{table*}

\section{Additional Experimental Results}
\label{EPIC:app:sec:experimental_results}

This section provides additional results 
for inference time, correctness, and ablation analyses.
We first report the full wall-clock inference times across all models, datasets, 
and denoising-step settings, complementing the relative inference-time results 
in Table~\ref{EPIC:main:tab:main_results}.
We then report syntactic and functional correctness, 
showing that \ours{} maintains correctness comparable to the prior CFG-constrained baseline 
while substantially reducing inference time.
Finally, we provide the full ablation results over all subsets of lexing
memoization, DFA-free validation, and relaxed compatible subset selection,
complementing the main ablation table in
Section~\ref{EPIC:main:ssec:ablation_results}.

\subsection{Full results on Inference Time}

Table~\ref{EPIC:app:tab:full_inference_time_results} provides 
the raw wall-clock inference times corresponding to the relative inference times 
reported in Table~\ref{EPIC:main:tab:main_results}.
The table includes unconstrained decoding, the prior CFG-constrained baseline,
and \ours{} for every model, task, and denoising-step setting.
These raw values are reported to make the normalization transparent 
and to show the absolute runtime scale across tasks.
As in the main results, 
all times are measured in seconds and averaged over three runtime seeds.

The raw times also show how the absolute decoding cost scales with the number
of denoising steps.
Although the increase is not exactly linear, 
runtime generally grows with the step count for all methods 
because each additional step requires another model forward pass.
This trend is visible even for unconstrained decoding, 
indicating that the overall runtime is still dominated 
by repeated model evaluation at larger step counts.
In this regime, \ours{} stays close to unconstrained decoding, 
whereas the baseline constrained decoder often remains noticeably slower 
due to additional per-step validity-checking overhead.
Thus, the absolute timings complement the relative inference-time results 
in the main text.
In many settings, \ours{} brings CFG-constrained decoding much closer 
to the runtime scale of unconstrained decoding.

\subsection{Full Results on Correctness}
\label{EPIC:app:ssec:correctness}
Tables~\ref{EPIC:app:tab:syntax_correctness} and
\ref{EPIC:app:tab:functional_correctness} report 
the full syntactic and functional correctness results.
We use U. for unconstrained decoding, C. for the prior CFG-constrained
baseline, and E. for \ours{}.
The superscript $-$ denotes the result 
before applying the final witness-based recovery procedure, 
while the version without $-$ includes this recovery.

\begin{table*}[t]
    \caption{
    Syntactic correctness across all models, denoising steps, and tasks.
    U. denotes unconstrained decoding, C. denotes the prior CFG-constrained
    baseline, and E. denotes \ours{}.
    The superscript $-$ reports correctness before witness-based recovery, while
    the version without $-$ includes recovery.
    All values are percentages.
    }
    \label{EPIC:app:tab:syntax_correctness}
    \centering
    \small
    \setlength{\tabcolsep}{3.0pt}
    \begin{tabular}{crrrrrrrrrrrrrrrr
    }
    \toprule[1.2pt]
        \multirow{2}{*}[-0.2em]{Model} & \multirow{2}{*}[-0.2em]{Steps} & 
        \multicolumn{5}{c}{C++} & \multicolumn{5}{c}{JSON} & \multicolumn{5}{c}{SMILES}\\
        \cmidrule(lr){3-7}
        \cmidrule(lr){8-12}
        \cmidrule(lr){13-17}
        & & U. & C.$^-$ & C. & E.$^-$ & \CCG E. & U. & C.$^-$ & C. & E.$^-$ & \CCG E. & U. & C.$^-$ & C. & E.$^-$ & \CCG E.\\
    \midrule[1.2pt]
        \multirow{5}{*}{Dream} & 16 &
        35.4 & 54.3 & \textbf{100} & 53.0 & \CCG \textbf{100} &
        32.4 & 50.4 & \textbf{100} & 50.0 & \CCG \textbf{100} &
        61.7 & 95.2 & \textbf{100} & 95.2 & \CCG \textbf{100} \\
                                  & 32 &
        67.7 & 80.5 & \textbf{100} & 80.5 & \CCG \textbf{100} &
        77.6 & 87.1 & \textbf{100} & 87.1 & \CCG \textbf{100} &
        70.7 & 98.2 & \textbf{100} & 98.2 & \CCG \textbf{100} \\
                                  & 64 &
        80.5 & 90.2 & \textbf{99.6} & 90.9 & \CCG \textbf{99.6} &
        92.3 & 95.6 & \textbf{100} & 96.0 & \CCG \textbf{100} &
        72.5 & 99.4 & \textbf{100} & 98.8 & \CCG \textbf{100} \\
                                  & 128 &
        89.0 & 95.1 & \textbf{100} & 95.1 & \CCG 99.8 &
        90.4 & 96.3 & \textbf{100} & 96.3 & \CCG \textbf{100} &
        73.1 & \textbf{100} & \textbf{100} & \textbf{100} & \CCG \textbf{100} \\
                                  & 256 &
        82.9 & 94.5 & \textbf{100} & 94.5 & \CCG \textbf{100} &
        91.2 & 96.1 & \textbf{100} & 96.3 & \CCG \textbf{100} &
        72.5 & \textbf{100} & \textbf{100} & \textbf{100} & \CCG \textbf{100} \\
    \midrule
        \multirow{5}{*}{DreamCoder} & 16 &
        17.7 & 33.5 & \textbf{99.8} & 39.0 & \CCG \textbf{99.8} &
        79.0 & 85.3 & \textbf{100} & 85.7 & \CCG \textbf{100} &
        78.4 & 97.0 & \textbf{100} & 97.0 & \CCG \textbf{100} \\
                                  & 32 &
        54.3 & 66.5 & \textbf{99.4} & 65.9 & \CCG \textbf{99.4} &
        87.5 & 91.9 & \textbf{100} & 92.3 & \CCG \textbf{100} &
        79.0 & 97.0 & \textbf{100} & 97.6 & \CCG \textbf{100} \\
                                  & 64 &
        72.6 & 77.4 & 99.2 & 78.7 & \CCG \textbf{100} &
        90.8 & 91.9 & \textbf{100} & 92.3 & \CCG \textbf{100} &
        74.9 & 98.8 & \textbf{100} & 98.8 & \CCG \textbf{100} \\
                                  & 128 &
        73.8 & 80.5 & \textbf{100} & 81.1 & \CCG \textbf{100} &
        93.0 & 94.1 & \textbf{100} & 94.1 & \CCG \textbf{100} &
        74.3 & 98.8 & \textbf{100} & 98.8 & \CCG \textbf{100} \\
                                  & 256 &
        84.1 & 85.0 & \textbf{100} & 86.0 & \CCG \textbf{100} &
        94.1 & 94.9 & \textbf{100} & 94.9 & \CCG \textbf{100} &
        74.3 & 98.8 & \textbf{100} & 98.8 & \CCG \textbf{100} \\
    \midrule
        \multirow{5}{*}{LLaDA} & 16 &
        5.5 & 34.1 & 99.6 & 32.9 & \CCG \textbf{100} &
        57.4 & 78.3 & \textbf{100} & 78.3 & \CCG \textbf{100} &
        41.3 & 90.4 & \textbf{100} & 89.2 & \CCG \textbf{100} \\
                                  & 32 &
        9.1 & 31.7 & 99.2 & 31.1 & \CCG \textbf{99.4} &
        75.4 & 89.0 & \textbf{100} & 88.6 & \CCG \textbf{100} &
        60.5 & 90.4 & \textbf{100} & 89.8 & \CCG \textbf{100} \\
                                  & 64 &
        42.7 & 67.1 & 99.8 & 67.1 & \CCG \textbf{100} &
        79.8 & 92.3 & \textbf{100} & 92.6 & \CCG \textbf{100} &
        65.9 & 94.0 & \textbf{100} & 93.4 & \CCG \textbf{100} \\
                                  & 128 &
        82.9 & 92.7 & \textbf{100} & 92.7 & \CCG \textbf{100} &
        87.9 & 96.3 & \textbf{100} & 96.3 & \CCG \textbf{100} &
        65.9 & 91.6 & \textbf{100} & 91.6 & \CCG \textbf{100} \\
                                  & 256 &
        96.3 & 99.4 & \textbf{100} & 99.4 & \CCG \textbf{100} &
        91.2 & 96.3 & \textbf{100} & 96.3 & \CCG \textbf{100} &
        65.3 & 95.8 & \textbf{100} & 95.6 & \CCG \textbf{100} \\
    \midrule
        \multirow{5}{*}{DiffuCoder} & 16 &
        25.0 & 40.9 & \textbf{99.8} & 39.0 & \CCG \textbf{99.8} &
        76.8 & 83.5 & \textbf{100} & 83.5 & \CCG \textbf{100} &
        70.7 & 94.0 & \textbf{100} & 93.4 & \CCG \textbf{100} \\
                                  & 32 &
        36.0 & 50.0 & \textbf{99.8} & 51.8 & \CCG \textbf{99.8} &
        87.9 & 90.8 & \textbf{100} & 90.4 & \CCG \textbf{100} &
        71.9 & 98.2 & \textbf{100} & 98.2 & \CCG \textbf{100} \\
                                  & 64 &
        53.7 & 67.1 & 99.8 & 67.7 & \CCG \textbf{100} &
        84.6 & 86.8 & \textbf{100} & 86.8 & \CCG \textbf{100} &
        62.3 & 98.8 & \textbf{100} & 98.8 & \CCG \textbf{100} \\
                                  & 128 &
        60.4 & 72.6 & \textbf{100} & 74.4 & \CCG 99.8 &
        77.9 & 79.0 & \textbf{100} & 79.0 & \CCG \textbf{100} &
        49.7 & 99.2 & \textbf{100} & 99.4 & \CCG \textbf{100} \\
                                  & 256 &
        93.9 & 91.3 & \textbf{100} & 93.7 & \CCG \textbf{100} &
        77.9 & 79.4 & \textbf{100} & 79.4 & \CCG \textbf{100} &
        47.9 & 97.0 & \textbf{100} & 97.0 & \CCG \textbf{100} \\
    \bottomrule[1.2pt]
    \end{tabular}
\end{table*}

\subsubsection{Syntactic Correctness}
Table~\ref{EPIC:app:tab:syntax_correctness} shows that both constrained
methods substantially improve syntactic correctness 
over unconstrained decoding.
Even before recovery, both C.$^-$ and E.$^-$ consistently improve over
unconstrained decoding, showing that rejection-based CFG checking already prevents 
syntactic errors.
\ours{} generally achieves pre-recovery syntactic correctness 
comparable to the prior CFG-constrained baseline.
The numbers are not always identical 
since \ours{} may commit several tokens in parallel, 
and therefore can fill positions in a different order from the sequential baseline.
This changes the subsequent partial outputs observed by the model, 
which can lead to differences in either direction.
In some settings, \ours{} is slightly higher than the baseline before recovery, 
while in others it is slightly lower.
These differences are expected 
because relaxed subset selection changes the commit order for efficiency, 
but every committed batch is still verified by the exact CFG checker.

After applying the recovery procedure, both constrained methods achieve
near-perfect syntactic correctness in almost all settings.
This confirms that the efficiency improvements of \ours{} do not weaken 
the final CFG-based acceptance criterion.
Rather, \ours{} mainly changes how compatible candidates are selected and
verified.

\begin{table*}[thb]
    \caption{
    Functional correctness across all models, denoising steps, and tasks.
    For C++, functional correctness is measured by passing the test cases.
    For JSON and SMILES, it is measured by task-specific normalized equivalence to
    the reference output.
    U. denotes unconstrained decoding, C. denotes the prior CFG-constrained
    baseline, and E. denotes \ours{}.
    The superscript $-$ reports results before witness-based recovery, while the
    version without $-$ includes recovery.
    All values are percentages.
    }
    \label{EPIC:app:tab:functional_correctness}
    \centering
    \small
    \setlength{\tabcolsep}{3.0pt}
    \begin{tabular}{crrrrrrrrrrrrrrrr
    }
    \toprule[1.2pt]
        \multirow{2}{*}[-0.2em]{Model} & \multirow{2}{*}[-0.2em]{Steps} & 
        \multicolumn{5}{c}{C++} & \multicolumn{5}{c}{JSON} & \multicolumn{5}{c}{SMILES}\\
        \cmidrule(lr){3-7}
        \cmidrule(lr){8-12}
        \cmidrule(lr){13-17}
        & & U. & C.$^-$ & C. & E.$^-$ & \CCG E. & U. & C.$^-$ & C. & E.$^-$ & \CCG E. & U. & C.$^-$ & C. & E.$^-$ & \CCG E.\\
    \midrule[1.2pt]
        \multirow{5}{*}{Dream} & 16 &
        10.4 & \textbf{11.6} & \textbf{11.6} & \textbf{11.6} & \CCG \textbf{11.6} &
        13.2 & 18.0 & \textbf{23.7} & 18.0 & \CCG \textbf{23.7} &
        \textbf{0.6} & \textbf{0.6} & \textbf{0.6} & \textbf{0.6} & \CCG \textbf{0.6} \\
                                  & 32 &
        15.2 & \textbf{16.5} & \textbf{16.5} & \textbf{16.5} & \CCG \textbf{16.5} &
        43.4 & 45.2 & \textbf{46.0} & 45.2 & \CCG \textbf{46.0} &
        \textbf{3.0} & \textbf{3.0} & \textbf{3.0} & \textbf{3.0} & \CCG \textbf{3.0} \\
                                  & 64 &
        18.1 & 18.5 & 18.5 & \textbf{18.9} & \CCG \textbf{18.9} &
        53.7 & \textbf{54.8} & \textbf{54.8} & \textbf{54.8} & \CCG \textbf{54.8} &
        3.0 & \textbf{3.6} & \textbf{3.6} & \textbf{3.6} & \CCG \textbf{3.6} \\
                                  & 128 &
        13.8 & \textbf{14.6} & \textbf{14.6} & 14.4 & \CCG 14.4 &
        51.5 & \textbf{53.3} & \textbf{53.3} & \textbf{53.3} & \CCG \textbf{53.3} &
        3.6 & \textbf{4.2} & \textbf{4.2} & \textbf{4.2} & \CCG \textbf{4.2} \\
                                  & 256 &
        12.8 & 13.4 & 13.4 & \textbf{14.0} & \CCG \textbf{14.0} &
        51.8 & \textbf{53.7} & \textbf{53.7} & \textbf{53.7} & \CCG \textbf{53.7} &
        3.0 & \textbf{4.2} & \textbf{4.2} & \textbf{4.2} & \CCG \textbf{4.2} \\
    \midrule
        \multirow{5}{*}{DreamCoder} & 16 &
        6.7 & 7.9 & 8.5 & \textbf{9.1} & \CCG \textbf{9.1} &
        41.9 & 43.8 & \textbf{44.1} & 43.4 & \CCG 44.0 &
        \textbf{4.2} & \textbf{4.2} & \textbf{4.2} & \textbf{4.2} & \CCG \textbf{4.2} \\
                                  & 32 &
        19.5 & 20.1 & 20.7 & 20.7 & \CCG \textbf{21.3} &
        52.2 & 52.6 & \textbf{53.3} & 52.6 & \CCG \textbf{53.3} &
        3.6 & \textbf{4.2} & \textbf{4.2} & \textbf{4.2} & \CCG \textbf{4.2} \\
                                  & 64 &
        26.8 & 26.8 & \textbf{27.2} & 26.2 & \CCG 26.6 &
        \textbf{53.7} & 53.3 & 53.3 & 53.3 & \CCG 53.3 &
        \textbf{2.4} & \textbf{2.4} & \textbf{2.4} & \textbf{2.4} & \CCG \textbf{2.4} \\
                                  & 128 &
        28.0 & \textbf{28.7} & \textbf{28.7} & \textbf{28.7} & \CCG \textbf{28.7} &
        55.9 & \textbf{56.6} & \textbf{56.6} & \textbf{56.6} & \CCG \textbf{56.6} &
        \textbf{2.4} & \textbf{2.4} & \textbf{2.4} & \textbf{2.4} & \CCG \textbf{2.4} \\
                                  & 256 &
        \textbf{28.0} & 27.4 & 27.4 & 27.4 & \CCG 27.4 &
        56.3 & \textbf{56.6} & \textbf{56.6} & \textbf{56.6} & \CCG \textbf{56.6} &
        \textbf{2.4} & \textbf{2.4} & \textbf{2.4} & \textbf{2.4} & \CCG \textbf{2.4} \\
    \midrule
        \multirow{5}{*}{LLaDA} & 16 &
        0.6 & 2.8 & 4.1 & 3.3 & \CCG \textbf{4.5} &
        39.0 & 43.8 & \textbf{44.1} & 43.8 & \CCG \textbf{44.1} &
        1.2 & \textbf{1.8} & \textbf{1.8} & \textbf{1.8} & \CCG \textbf{1.8} \\
                                  & 32 &
        3.0 & 5.1 & 5.3 & 5.5 & \CCG \textbf{6.3} &
        44.1 & 49.6 & \textbf{50.0} & 49.6 & \CCG \textbf{50.0} &
        0.6 & \textbf{1.2} & \textbf{1.2} & \textbf{1.2} & \CCG \textbf{1.2} \\
                                  & 64 &
        10.4 & \textbf{11.0} & \textbf{11.0} & \textbf{11.0} & \CCG \textbf{11.0} &
        44.5 & 49.6 & \textbf{50.0} & 49.6 & \CCG \textbf{50.0} &
        1.2 & \textbf{1.8} & \textbf{1.8} & \textbf{1.8} & \CCG \textbf{1.8} \\
                                  & 128 &
        \textbf{15.9} & \textbf{15.9} & \textbf{15.9} & \textbf{15.9} & \CCG \textbf{15.9} &
        48.2 & \textbf{52.9} & \textbf{52.9} & \textbf{52.9} & \CCG \textbf{52.9} &
        \textbf{1.8} & \textbf{1.8} & \textbf{1.8} & \textbf{1.8} & \CCG \textbf{1.8} \\
                                  & 256 &
        \textbf{25.6} & 25.0 & 25.0 & 25.0 & \CCG 25.0 &
        49.6 & 52.2 & \textbf{52.9} & 52.2 & \CCG \textbf{52.9} &
        \textbf{2.4} & \textbf{2.4} & \textbf{2.4} & \textbf{2.4} & \CCG \textbf{2.4} \\
    \midrule
        \multirow{5}{*}{DiffuCoder} & 16 &
        12.8 & 14.0 & 14.0 & \textbf{14.6} & \CCG \textbf{14.6} &
        42.3 & 43.8 & \textbf{44.1} & 43.8 & \CCG \textbf{44.1} &
        0.0 & \textbf{0.6} & \textbf{0.6} & \textbf{0.6} & \CCG \textbf{0.6} \\
                                  & 32 &
        15.2 & \textbf{18.3} & \textbf{18.3} & 17.7 & \CCG 17.7 &
        47.8 & 48.9 & \textbf{49.6} & 48.5 & \CCG 49.3 &
        0.0 & 0.0 & 0.0 & 0.0 & \CCG 0.0 \\
                                  & 64 &
        17.7 & \textbf{18.3} & \textbf{18.3} & \textbf{18.3} & \CCG \textbf{18.3} &
        45.2 & 47.1 & \textbf{47.4} & 47.1 & \CCG \textbf{47.4} &
        \textbf{0.6} & \textbf{0.6} & \textbf{0.6} & \textbf{0.6} & \CCG \textbf{0.6} \\
                                  & 128 &
        25.6 & 27.4 & \textbf{29.9} & 27.4 & \CCG 29.7 &
        42.3 & 43.4 & \textbf{43.8} & 43.4 & \CCG \textbf{43.8} &
        \textbf{1.8} & \textbf{1.8} & \textbf{1.8} & \textbf{1.8} & \CCG \textbf{1.8} \\
                                  & 256 &
        43.9 & 44.3 & 45.5 & 44.5 & \CCG \textbf{45.7} &
        41.9 & \textbf{43.4} & \textbf{43.4} & \textbf{43.4} & \CCG \textbf{43.4} &
        \textbf{2.4} & \textbf{2.4} & \textbf{2.4} & \textbf{2.4} & \CCG \textbf{2.4} \\
    \bottomrule[1.2pt]
    \end{tabular}
\end{table*}

\subsubsection{Functional Correctness}
Table~\ref{EPIC:app:tab:functional_correctness} reports task-level functional
correctness.
Constrained decoding often improves functional correctness over unconstrained decoding, 
especially on C++ and JSON, since syntactically invalid outputs
cannot satisfy the downstream evaluator.
As the number of denoising steps increases, 
functional correctness also tends to improve.
This is consistent with the usual diffusion decoding trade-off.
With more steps, the decoder unmasks fewer tokens at each step, 
making each update more conditioned on previously committed context 
and generally improving generation quality.

However, the improvement in syntactic correctness does not always translate
into a proportional improvement in functional correctness.
This behavior is also observed in prior work and reflects a common limitation
of grammar-only constrained decoding.
A grammar can enforce formal syntactic validity, 
but it does not verify task semantics.
The witness-based recovery procedure has only a marginal effect 
on functional correctness in most settings.
Although recovery raises syntactic correctness to nearly 100\%, 
the recovered completion is selected to satisfy the grammar 
rather than to optimize the task-level objective.
Therefore, recovery is useful as a syntactic fallback, but it should not be
expected to substantially improve semantic or functional quality.

The gap between syntactic and functional correctness 
is particularly visible on SMILES.
Although the grammar can enforce the surface syntax of SMILES strings, 
it does not fully characterize task-level molecular correctness 
or equivalence to the target molecule.
In addition, the underlying CFG is only an approximation 
to the full set of functionally valid molecular strings.
As a result, syntactic correctness can approach 100\% 
while functional correctness remains low.

\subsection{Full Ablation Results}
\label{EPIC:app:ssec:ablation_results}

Table~\ref{EPIC:app:tab:full_ablation_results} reports the full ablation results 
at $32$ denoising steps.
The table includes all individual components, all pairwise combinations, 
and the full \ours{} configuration.
Overall, the results show that the three components provide complementary sources of speedup.
Lexing memoization mainly reduces repeated lexical analysis, 
DFA-free validation removes the exact-checking dependence on repeated DFA construction,
and relaxed compatible subset selection reduces the number of sequential 
exact checks by recovering parallel commitment.

The effect of each component depends on the task and model.
On C++, all three components are consistently useful 
because the grammar and lexical specification are more complex, 
making repeated lexing, automaton construction, and CFG validation expensive.
DFA-free validation is particularly effective for DreamCoder and DiffuCoder on C++, 
while relaxed subset selection gives large gains for LLaDA, 
whose block-wise decoding schedule provides stable candidate batches.
On JSON, lexing memoization and relaxed subset selection are generally more
effective than DFA-free validation alone.
This is consistent with the fact that JSON validation is relatively cheap 
once the partial-output representation has been constructed, 
so avoiding DFA construction alone does not always compensate 
for the overhead of graph parsing.
On SMILES, the differences are small 
because the baseline constraint overhead is already low. 
In some cases, the added component can slightly increase runtime.

Pairwise combinations usually improve over the corresponding single-component variants, 
indicating that the optimizations address different parts of the pipeline.
For example, combining memoization with DFA-free validation substantially reduces C++ runtime, 
while combining memoization with relaxed subset selection is especially helpful on JSON.
The full configuration, which enables all three components, gives the best runtime 
in most settings and is never substantially worse than the best ablation.
These results support the design of \ours{} as a combination of complementary
optimizations rather than a speedup from a single dominant component.

\begin{table}[t]
    \caption{
        Full ablation results for \ours{}.
        Starting from the prior CFG-constrained decoder (\textsc{Con.}),
        we separately add $\circled{1}$ lexing memoization,
        $\circled{2}$ DFA-free validation with Earley-style parsing,
        and $\circled{3}$ relaxed compatible subset selection for parallel commit.
        \ours{} enables all three components.
    }
    \label{EPIC:app:tab:full_ablation_results}
    \centering
    \small
    \begin{tabular}{clrrrr
    }
    \toprule[1.2pt]
        Model & Method & ~~~~C++ & ~~JSON & SMILES\\
    \midrule[1.2pt]
        \multirow{8}{*}{Dream} & Con. &
        900 & 1203 & \textbf{516} \\
                                  & \circled{1} &
        843 & 1095 & 541 \\
                                  & \circled{2} &
        863 & 1237 & 540 \\
                                  & \circled{3} &
        900 & 1162 & 545 \\
                                  &\circled{1}+\circled{2}&
        758 & 1071 & 540 \\
                                  &\circled{1}+\circled{3}&
        795 & 1043 & 545 \\
                                  &\circled{2}+\circled{3}&
        844 & 1160 & 545 \\
                                  &\CCG \circled{1}+\circled{2}+\circled{3}&
        \CCG \textbf{700} & \CCG \textbf{983} & \CCG 519 \\
    \midrule
        \multirow{8}{*}{DreamCoder} & Con. &
        1582 & 1089 & \textbf{523} \\
                                  & \circled{1} &
        1385 & 939 & 540 \\
                                  & \circled{2} &
        1261 & 1109 & 540 \\
                                  & \circled{3} &
        1378 & 986 & 543 \\
                                  &\circled{1}+\circled{2}&
        1006 & 907 & 540 \\
                                  &\circled{1}+\circled{3}&
        1056 & 774 & 543 \\
                                  &\circled{2}+\circled{3}&
        1154 & 974 & 543 \\
                                  &\CCG \circled{1}+\circled{2}+\circled{3}&
        \CCG \textbf{824} & \CCG \textbf{735} & \CCG \textbf{523} \\
    \midrule
        \multirow{8}{*}{LLaDA} & Con. &
        791 & 1554 & 636 \\
                                  & \circled{1} &
        772 & 1450 & 655 \\
                                  & \circled{2} &
        738 & 1619 & 628 \\
                                  & \circled{3} &
        601 & 1179 & 609 \\
                                  &\circled{1}+\circled{2}&
        704 & 1442 & 627 \\
                                  &\circled{1}+\circled{3}&
        577 & 1129 & 608 \\
                                  &\circled{2}+\circled{3}&
        571 & 1176 & 595 \\
                                  &\CCG \circled{1}+\circled{2}+\circled{3}&
        \CCG \textbf{531} & \CCG \textbf{1076} & \CCG \textbf{570} \\
    \midrule
        \multirow{8}{*}{DiffuCoder} & Con. &
        1278 & 1314 & 584 \\
                                  & \circled{1} &
        1152 & 1172 & 598 \\
                                  & \circled{2} &
        1085 & 1307 & 582 \\
                                  & \circled{3} &
        1119 & 1195 & 601 \\
                                  &\circled{1}+\circled{2}&
        918 & 1129 & 581 \\
                                  &\circled{1}+\circled{3}&
        947 & 1054 & 599 \\
                                  &\circled{2}+\circled{3}&
        996 & 1174 & 597 \\
                                  &\CCG \circled{1}+\circled{2}+\circled{3}&
        \CCG \textbf{862} & \CCG \textbf{1005} & \CCG \textbf{576} \\
    \bottomrule[1.2pt]
    \end{tabular}
\end{table}

\subsection{DFA-Free Validator Analysis}
\label{EPIC:app:ssec:validator_analysis}

We further analyze the cost of the DFA-free exact validator independently
of model inference.
For a lexeme-path graph with $N$ nodes and
\[
R=\sum_{A\to\alpha\in P} (|\alpha|+1)
\]
dotted grammar positions, the number of distinct parser items is bounded by
$O(RN^2)$.
Each item is identified by its grammar position, origin graph node, and
current graph node, and duplicate items are removed before being added to
the worklist.
This bound characterizes the parser state space rather than the overall
worst-case parsing complexity, which remains comparable to general CFG
recognition.

Table~\ref{EPIC:app:tab:validator_schema_complexity} reports per-check
runtime as the grammar size increases.
The total cost includes construction of the lexeme-path graph and exact
graph-based CFG validation.
Validation becomes more expensive for larger grammars and larger graphs,
but the exact parser itself remains a small fraction of the total
validator-side cost.
Even for the largest grammar-size bin, exact graph parsing takes
$0.124$ ms per check on average, compared with $1.316$ ms for graph
construction.

\begin{table}[t]
    \caption{
    Per-check cost of DFA-free validation across JSON schema grammars
    with different numbers of production rules.
    Total time includes lexeme-path graph construction and exact graph parsing.
    }
    \label{EPIC:app:tab:validator_schema_complexity}
    \centering
    \small
    \begin{tabular}{lrcccr}
    \toprule[1.2pt]
    CFG size &
    \# Checks &
    Total~(ms) &
    Parsing~(ms) &
    Graph~(ms) &
    \# States \\
    \midrule[1.2pt]
    $<30$ rules
        & 10,129 & 0.728 & 0.039 & 0.686 & 19.5 \\
    $30$--$59$ rules
        & 11,397 & 0.858 & 0.053 & 0.802 & 26.8 \\
    $>59$ rules
        & 5,079 & 1.442 & 0.124 & 1.316 & 224.3 \\
    \bottomrule[1.2pt]
    \end{tabular}
\end{table}

Mask density shows a different pattern.
As shown in Table~\ref{EPIC:app:tab:validator_mask_density},
validation cost is not monotonic in the fraction of masked positions.
Highly masked sequences contain little fixed text, whereas intermediate
mask densities can induce more fragmented fixed regions and larger
lexeme-path graphs.
This suggests that graph structure and lexical ambiguity, rather than mask
density alone, determine the practical cost of exact validation.

\begin{table}[t]
    \caption{
    Per-check cost of DFA-free validation at different mask densities,
    defined as the fraction of masked positions in the current partial output.
    Total time includes lexeme-path graph construction and exact graph parsing.
    }
    \label{EPIC:app:tab:validator_mask_density}
    \centering
    \small
    \setlength{\tabcolsep}{4pt}
    \begin{tabular}{lrcccr}
    \toprule[1.2pt]
    Mask density &
    \# Checks &
    Total~(ms) &
    Parsing~(ms) &
    Graph~(ms) &
    \# States \\
    \midrule[1.2pt]
    $[0,.25)$
        & 16,233 & 0.891 & 0.058 & 0.831 & 56.3 \\
    $[.25,.50)$
        & 7,079 & 0.980 & 0.068 & 0.909 & 84.3 \\
    $[.50,.75)$
        & 2,240 & 1.011 & 0.067 & 0.941 & 51.8 \\
    $[.75,1]$
        & 1,053 & 0.755 & 0.056 & 0.696 & 14.3 \\
    \bottomrule[1.2pt]
    \end{tabular}
\end{table}

\subsection{Regular Cover Analysis}
\label{EPIC:app:ssec:regular_cover_analysis}

We next characterize the preprocessing cost and operational tightness of
the regular cover used for parallel subset selection.
The cover is constructed by flattening the recursive transition network
induced by the CFG into an ENFA.
This cover is determinized and minimized once for each grammar and then cached.
Unlike the baseline validity checker, this conversion is not repeated for
individual candidate updates.
Table~\ref{EPIC:app:tab:cover_construction} reports the construction cost and
automaton size for the instance-specific JSON schema grammars.
The size of the intermediate ENFA increases with grammar complexity, while
the minimized cover DFA remains small across all schema-size bins.
Construction is also inexpensive, taking approximately $0.4$--$1.3$ ms on
average.

\begin{table}[ht]
    \caption{
    Construction cost and automaton size of regular covers for
    instance-specific JSON schema grammars.
    Build time includes cover construction, determinization, and minimization.
    }
    \label{EPIC:app:tab:cover_construction}
    \centering
    \small
    \setlength{\tabcolsep}{4pt}
    \begin{tabular}{lrcccc}
    \toprule[1.2pt]
    CFG size &
    \# Unique Grammars &
    \# Production Rules &
    \# ENFA states &
    \# DFA states &
    Build Time~(ms) \\
    \midrule[1.2pt]
    $<30$ rules
        & 26  & 25.8 & 107.3 & 8.5 & 0.432 \\
    $30$--$59$ rules
        & 143 & 36.2 & 147.1 & 9.7 & 0.574 \\
    $>59$ rules
        & 68  & 79.4 & 318.3 & 8.0 & 1.334 \\
    \bottomrule[1.2pt]
    \end{tabular}
\end{table}

We further examine how closely the regular cover approximates the CFG 
during decoding.
Since the cover defines a language superset, a cover-compatible subset may
still be rejected by the subsequent exact CFG check.
We measure this operational tightness using the false-positive rate of
initially selected cover-compatible subsets.
Table~\ref{EPIC:app:tab:cover_tightness} groups JSON instances according to
whether any such false positive is observed during decoding.
Most instances exhibit no false positives.
For instances where false positives occur, the average parallel commit size
decreases from $2.73$ to $1.53$.
This indicates that a looser cover primarily reduces the amount of parallelism
recovered by subset selection, while correctness is preserved by the
subsequent exact CFG verification.

\begin{table}[t]
    \caption{
    Operational tightness of the regular cover on JSON.
    FP rate denotes the fraction of initially selected cover-compatible subsets
    rejected by the subsequent exact CFG check.
    Average parallel commit size is measured after exact verification.
    Instances for which no nontrivial cover-compatible subset is found are omitted.
    }
    \label{EPIC:app:tab:cover_tightness}
    \centering
    \small
    \setlength{\tabcolsep}{5pt}
    \begin{tabular}{lrrr}
    \toprule[1.2pt]
    Instance Group &
    \# Inst. &
    FP Rate &
    Parallel Commit \\
    \midrule[1.2pt]
    No FP observed
        & 244 & 0.0\%  & 2.73 \\
    FP observed
        & 22  & 50.5\% & 1.53 \\
    \bottomrule[1.2pt]
    \end{tabular}
\end{table}

\subsection{Output Fidelity to the Constrained Baseline}
\label{EPIC:app:ssec:output_fidelity}

Although \ours{} preserves the same CFG-completability criterion as the
baseline, parallel commitment can change the order in which partial outputs
are presented to the model and may therefore alter the generated sequence.
We further examine this effect by comparing the outputs of \ours{} and the
prior constrained baseline on C++.
Table~\ref{EPIC:app:tab:output_fidelity} reports functional accuracy,
overlap between the sets of correctly solved instances, and exact output agreement.
Let $C_{\mathrm{base}}$ and $C_{\ours{}}$ denote the sets of instances
that are functionally correct under the baseline and \ours{}, respectively.
We compute their Jaccard similarity as
\[
J(C_{\mathrm{base}}, C_{\ours{}})
=
\frac{|C_{\mathrm{base}} \cap C_{\ours{}}|}
{|C_{\mathrm{base}} \cup C_{\ours{}}|}.
\]
The mean functional accuracy remains nearly unchanged, from $18.72\%$ for
the baseline to $18.86\%$ for \ours{}.
The correct-instance sets have a Jaccard similarity of $94.20\%$, and
$97.58\%$ of the instances solved by the baseline are also solved by
\ours{}.
In addition, $73.13\%$ of generated outputs exactly match the baseline text.
These results indicate that parallel commitment largely preserves the
functional behavior of the original constrained decoder 
while reducing its decoding cost.

\begin{table}[ht]
    \caption{
    Output fidelity of \ours{} relative to the prior constrained baseline
    on C++.
    Correct-set Jaccard measures the overlap between the sets of functionally
    correct instances produced by the two methods.
    }
    \label{EPIC:app:tab:output_fidelity}
    \centering
    \small
    \begin{tabular}{lr}
    \toprule[1.2pt]
    Metric & Value \\
    \midrule[1.2pt]
    Baseline Functional Accuracy & 18.72\% \\
    \ours{} Functional Accuracy & 18.86\% \\
    Correct-set Jaccard & 94.20\% \\
    Baseline-to-\ours{} Correct Overlap & 97.58\% \\
    \ours-to-Baseline Correct Overlap & 95.86\% \\
    Exact Text Match & 73.13\% \\
    \bottomrule[1.2pt]
    \end{tabular}
\end{table}

\begin{figure*}[t]
    \centering
    \includegraphics[width=\textwidth]{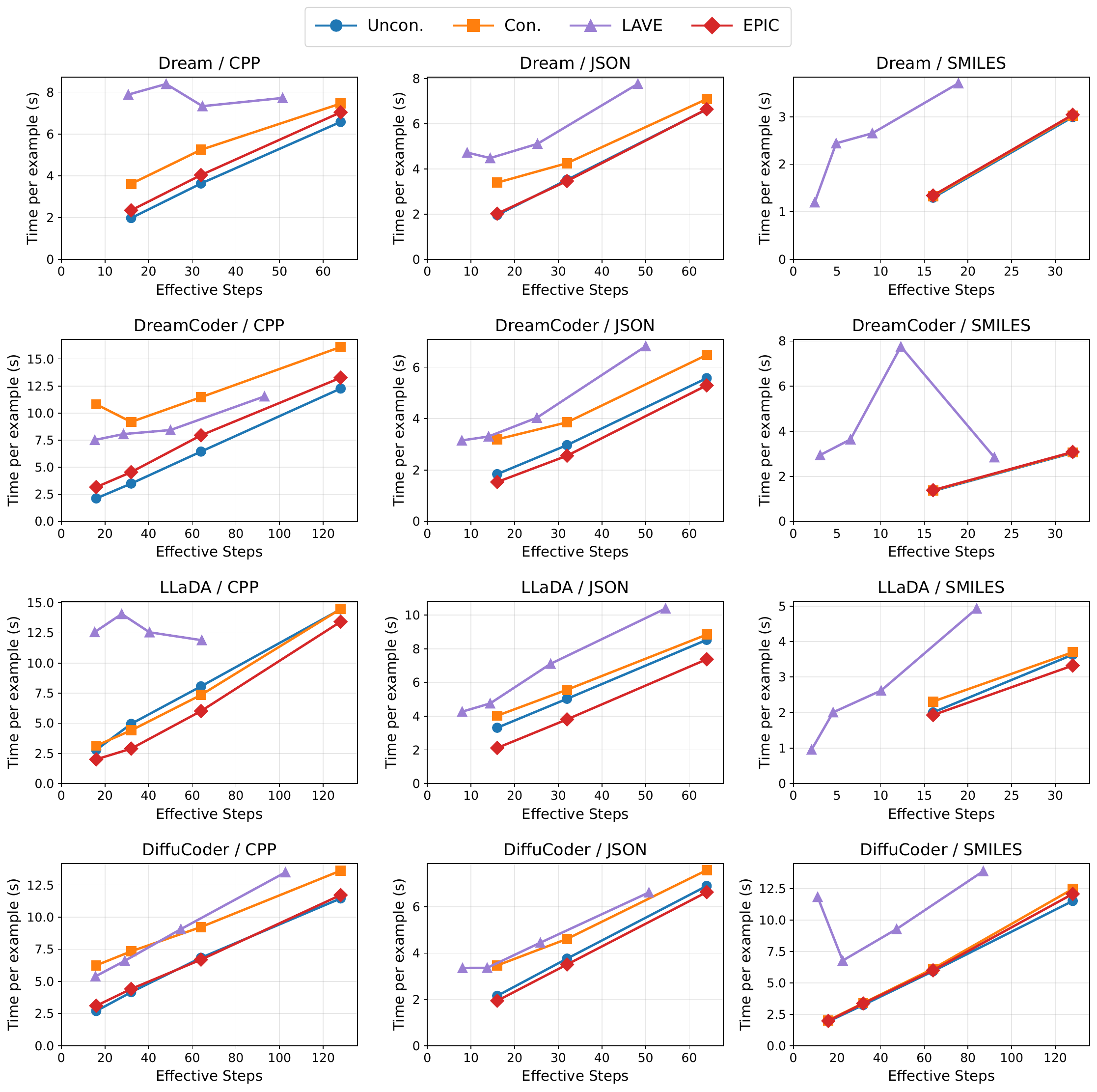}
    \caption{
    Runtime of unconstrained decoding, Con., LAVE, and \ours{}
    across effective decoding steps.
    }
    \label{EPIC:app:fig:lave_runtime}
\end{figure*}

\subsection{Comparison with LAVE}
\label{EPIC:app:ssec:lave_comparison}

We additionally compare \ours{} with LAVE~\citep{ZhangLLLJLL2026},
a recent CFG-constrained decoding method for DLMs.
We report this comparison separately from the main results because
LAVE uses a different decoding protocol.
In particular, LAVE adopts semi-autoregressive decoding 
with a fixed block size of $32$ for all models, 
rather than preserving the native decoding configuration of each model.
Using smaller blocks limits the amount of masked context considered 
within each block 
and thereby also helps reduce the cost of repeated lookahead verification.
In contrast, our main experiments follow the decoding schedules 
used by the respective models and the prior completion-based CFG decoder.
Consequently, the same nominal number of denoising steps does not necessarily
correspond to the same amount of generation progress across the two protocols,
making a direct comparison in the main tables potentially misleading.

To provide a comparison under conditions that are as similar as possible,
we integrate LAVE into the same evaluation pipeline 
and match the model checkpoints, tokenizers, datasets, prompts, 
maximum generation length, seeds, hardware, and evaluators.
We retain LAVE's semi-autoregressive decoding procedure with a block size
of $32$, temperature $0.2$, $N=10$ lookahead samples, 
and a retry budget of $\tau=5$.
Since LAVE generates the output semi-autoregressively in blocks,
its nominal step budget is not directly comparable to 
the sequence-level step budgets used for Uncon., Con., and \ours{}.
In particular, LAVE may terminate before visiting all blocks, 
in which case part of its nominal step budget is never used.
We therefore compare the methods using an upper bound 
on the number of executed denoising steps.
For Uncon., Con., and \ours{}, this is simply the configured number of steps.
For LAVE, we estimate this upper bound as the average number of visited blocks
multiplied by the configured number of steps per block.

\begin{figure*}[t]
    \centering
    \includegraphics[width=\textwidth]{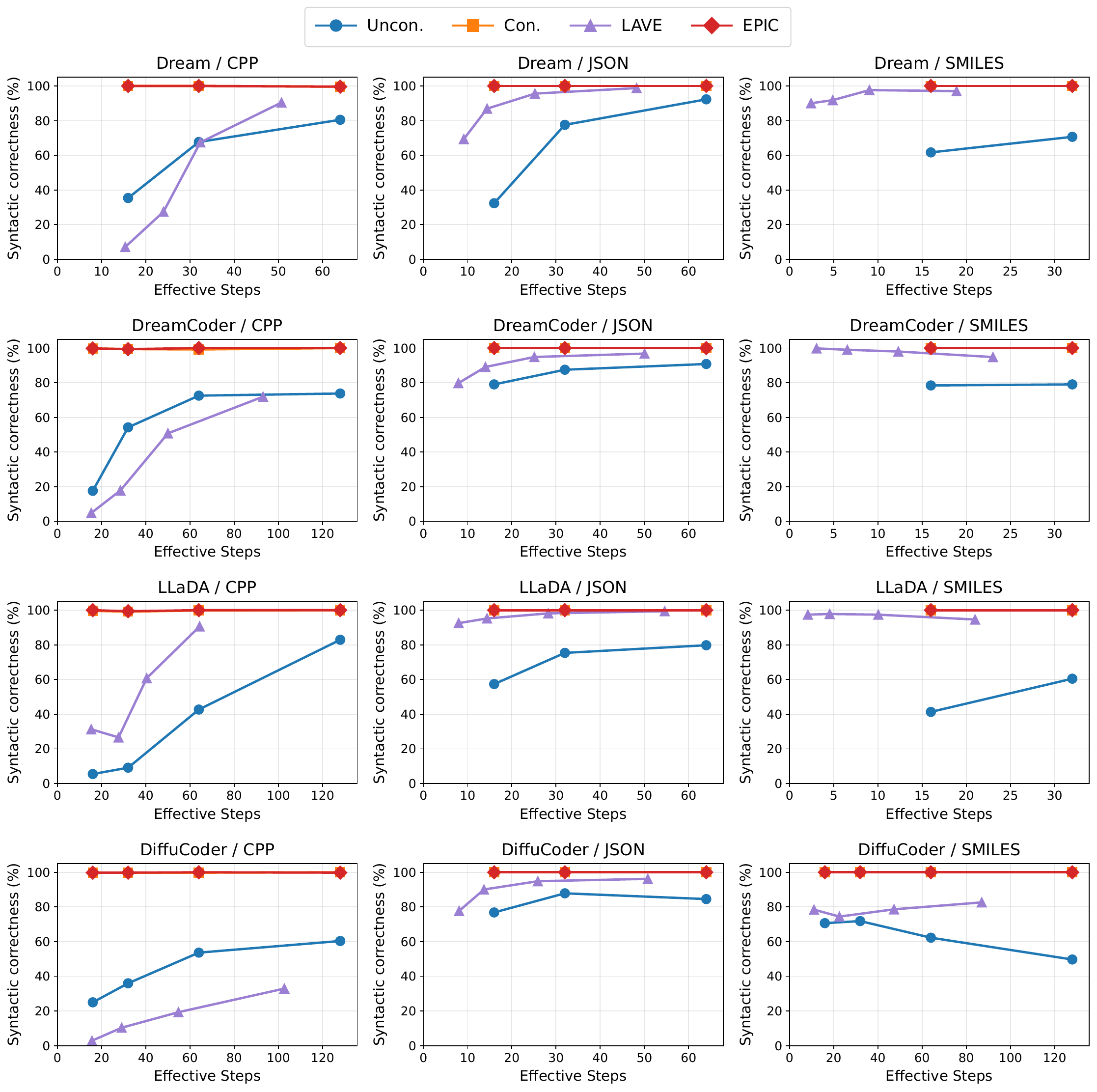}
    \caption{
    Syntactic correctness of unconstrained decoding, Con., LAVE, and \ours{}
    across effective decoding steps.
    }
    \label{EPIC:app:fig:lave_syntactic}
\end{figure*}

Figure~\ref{EPIC:app:fig:lave_runtime} compares wall-clock runtime at
comparable effective decoding progress.
Across most model and task combinations, \ours{} remains close to
unconstrained decoding and is substantially faster than LAVE.
The difference is particularly visible on C++ and JSON, where repeated
lookahead verification introduces a larger runtime cost.
On SMILES, where the grammar is substantially smaller, the difference is
less consistent and there is less room for either constrained decoder 
to reduce overhead relative to the underlying generation cost.

\begin{figure*}[t]
    \centering
    \includegraphics[width=\textwidth]{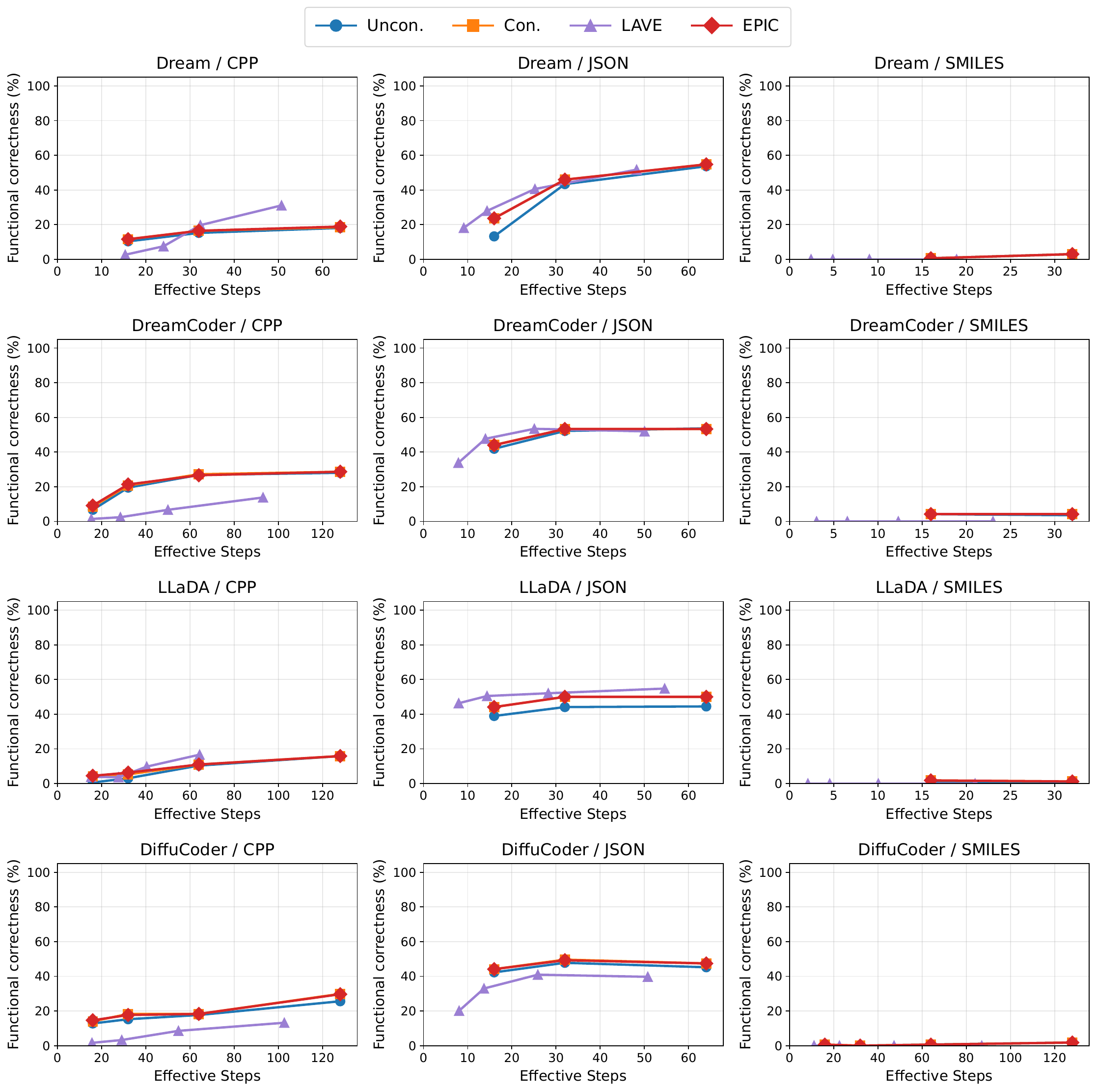}
    \caption{
    Functional correctness of unconstrained decoding, Con., LAVE, and \ours{}
    across effective decoding steps.
    }
    \label{EPIC:app:fig:lave_functional}
\end{figure*}

Figures~\ref{EPIC:app:fig:lave_syntactic} and
\ref{EPIC:app:fig:lave_functional} compare correctness under the same view.
\ours{} maintains near-perfect syntactic correctness over a broad range of
effective steps, while LAVE is more sensitive to decoding progress,
particularly on C++.
For functional correctness, the two methods are comparable in some
settings, but LAVE does not show a consistent advantage despite its larger
runtime.
Overall, these results support the same efficiency-correctness trade-off
observed in the main experiments. \ours{} substantially reduces constrained
decoding cost while maintaining strong syntactic and functional correctness.

\end{document}